%% file: main.tex
\documentclass{article}

\usepackage{microtype}
\usepackage{graphicx}
\usepackage{subfigure}
\usepackage{booktabs} % for professional tables

\usepackage{hyperref}

\usepackage[accepted]{icml2024}

\usepackage{amsmath}
\usepackage{amssymb}
\usepackage{mathtools}
\usepackage{amsthm}

\usepackage[capitalize,noabbrev]{cleveref}

\theoremstyle{plain}

\theoremstyle{definition}

\theoremstyle{remark}

\usepackage[textsize=tiny]{todonotes}
\usepackage{color,xcolor}

\newcommand{\method}{\text{aL\small{LM}4T\small{S}}}

\newcommand{\boldres}[1]{{\textbf{\textcolor{red}{#1}}}}
\newcommand{\secondres}[1]{{\underline{#1}}}
\newcommand*{\shortautoref}[1]{%
  \begingroup
    \def\sectionautorefname{Sec.}%
    \def\subsectionautorefname{Sec.}%
    \def\figureautorefname{Fig.}%
    \def\tableautorefname{Tab.}%
    \def\equationautorefname{Eq.}%
    \autoref{#1}%
  \endgroup
}
\usepackage{multirow}
\usepackage{pifont}
\usepackage{makecell}
\usepackage{threeparttable}
\usepackage{tablefootnote}
\input{math_commands.tex}

\icmltitlerunning{Multi-Patch Prediction: Adapting LLMs for Time Series Representation Learning}

\begin{document}

\twocolumn[
\icmltitle{Multi-Patch Prediction: Adapting LLMs for Time Series \texorpdfstring{\\}{} Representation Learning}



\icmlsetsymbol{equal}{*}

\begin{icmlauthorlist}
\icmlauthor{Yuxuan Bian}{equal,cuhk,tj}
\icmlauthor{Xuan Ju}{equal,cuhk}
\icmlauthor{Jiangtong Li}{equal,tj,sjtu}
\icmlauthor{Zhijian Xu}{cuhk}
\icmlauthor{Dawei Cheng}{tj,ailab}
\icmlauthor{Qiang Xu}{cuhk}
% \icmlauthor{Firstname3 Lastname3}{comp}
\end{icmlauthorlist}

\icmlaffiliation{cuhk}{The Chinese University of Hong Kong}
\icmlaffiliation{tj}{Tongji University}
\icmlaffiliation{sjtu}{Shanghai Jiao Tong University}
\icmlaffiliation{ailab}{Shanghai Artificial Intelligence Laboratory}
 
% \icmlaffiliation{tj}{Department of Computer Science and Technology, Tongji University, Shanghai, China}
% \icmlaffiliation{cuhk}{Department of Computer Science and Technology, The Chinese University of Hong Kong, Hong Kong SAR, China}

\icmlcorrespondingauthor{Dawei Cheng}{dcheng@tongji.edu.cn}
\icmlcorrespondingauthor{Qiang Xu}{qxu@cse.cuhk.edu.hk}

\icmlkeywords{Time Series Analysis, Time Series Representation Learning, Large Language Models}
\vskip 0.3in
]



\printAffiliationsAndNotice{\icmlEqualContribution} % otherwise use the standard text.

\begin{abstract}
In this study, we present \emph{\method{}}, an innovative framework that adapts Large Language Models (LLMs) for time-series representation learning. Central to our approach is that we reconceive time-series forecasting as a self-supervised, multi-patch prediction task, which, compared to traditional contrastive learning or mask-and-reconstruction methods, captures temporal dynamics in patch representations more effectively. Our strategy encompasses two-stage training: (i). a causal continual pre-training phase on various time-series datasets, anchored on next patch prediction, effectively syncing LLM capabilities with the intricacies of time-series data; (ii). fine-tuning for multi-patch prediction in the targeted time-series context. A distinctive element of our framework is the patch-wise decoding layer, which departs from previous methods reliant on sequence-level decoding. Such a design directly transposes individual patches into temporal sequences, thereby significantly bolstering the model's proficiency in mastering temporal patch-based representations. \method{} demonstrates superior performance in several downstream tasks, proving its effectiveness in deriving temporal representations with enhanced transferability and marking a pivotal advancement in the adaptation of LLMs for time-series analysis.
\end{abstract}

\input{sections/1-introduction}
\input{sections/2-related_work}
\input{sections/3-preliminaries_and_motivation}
\input{sections/4-method}
\input{sections/5-experiments}
\input{sections/6-conclusion}

\section*{IMPACT STATEMENTS}
This paper presents work whose goal is to advance the field of Machine Learning. There are many potential societal consequences of our work, none of which we feel must be specifically highlighted here.

\bibliography{main.bib}
\bibliographystyle{icml2024}

%%%%%%%%%%%%%%%%%%%%%%%%%%%%%%%%%%%%%%%%%%%%%%%%%%%%%%%%%%%%%%%%%%%%%%%%%%%%%%%
%%%%%%%%%%%%%%%%%%%%%%%%%%%%%%%%%%%%%%%%%%%%%%%%%%%%%%%%%%%%%%%%%%%%%%%%%%%%%%%
% APPENDIX
%%%%%%%%%%%%%%%%%%%%%%%%%%%%%%%%%%%%%%%%%%%%%%%%%%%%%%%%%%%%%%%%%%%%%%%%%%%%%%%
%%%%%%%%%%%%%%%%%%%%%%%%%%%%%%%%%%%%%%%%%%%%%%%%%%%%%%%%%%%%%%%%%%%%%%%%%%%%%%%
\newpage

\input{sections/7-appendix}

%%%%%%%%%%%%%%%%%%%%%%%%%%%%%%%%%%%%%%%%%%%%%%%%%%%%%%%%%%%%%%%%%%%%%%%%%%%%%%%
%%%%%%%%%%%%%%%%%%%%%%%%%%%%%%%%%%%%%%%%%%%%%%%%%%%%%%%%%%%%%%%%%%%%%%%%%%%%%%%

\end{document}

%% file: math_commands.tex
\usepackage{amsmath,amsfonts,bm}

\def\eqref#1{equation~\ref{#1}}
\def\1{\bm{1}}

\def\vp{{\bm{p}}}

\def\vx{{\bm{x}}}

\DeclareMathAlphabet{\mathsfit}{\encodingdefault}{\sfdefault}{m}{sl}
\SetMathAlphabet{\mathsfit}{bold}{\encodingdefault}{\sfdefault}{bx}{n}

\newcommand{\E}{\mathbb{E}}

\newcommand{\R}{\mathbb{R}}

%% file: sections/1-introduction.tex
\vspace{-5.5mm}
\section{INTRODUCTION}\label{sec-introduction}
Time-series analysis (TSA) plays a pivotal role in a myriad of real-world applications. Current state-of-the-art TSA methodologies are usually custom-designed for specific tasks, such as forecasting \citep{DLinear,PatchTST}, classification \citep{dempster2020rocket}, and anomaly detection \citep{xu2021anomaly}. 

Despite these advancements, the quest for a versatile time-series representation capable of addressing diverse downstream tasks remains a formidable challenge. Traditional approaches predominantly rely on self-supervised learning strategies such as contrastive learning \citep{ts2vec,btsf} and mask-and-reconstruction modeling \citep{zerveas2021transformer,Ti-MAE}. 
% Yet, these techniques often struggle to fully grasp the intricate temporal variations characteristic of time-series data~\citep{bad-contrasive-learning}.
Yet, they often struggle to fully grasp the intricate temporal variations characteristic of time-series, arising from inconsistencies between high-level representation
optimization with downstream low-level tasks~\citep{bad-contrasive-learning}, or the temporal disruption caused by random masking~\citep{ma2023survey}.

\begin{figure*}[!t]
  \centering
  \vspace{-2.5mm}
  \includegraphics[width=1.\textwidth]{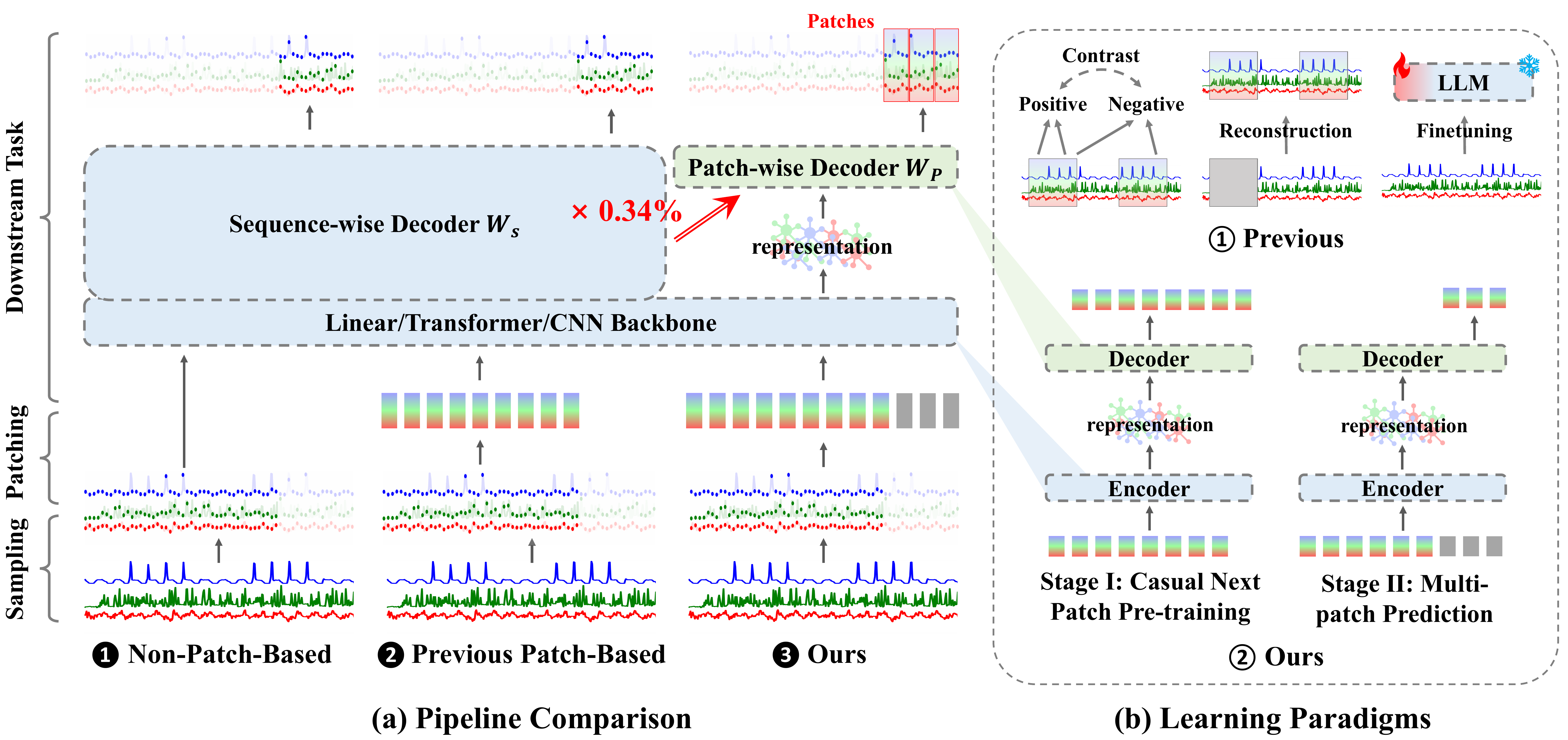}
  \vspace{-8mm}
  \caption{
  \textbf{Pipeline Comparison.} Given a time series embedding/patch sequence $\vx \in \R^{L \times D}, D \gg P$ where $P$ is the patch size and forecasting horizon $H$:
  Non-Patch Based Models~\ding{182} or Patch Based Models~\ding{183} map it to the target sequence using a huge sequence-level linear layer $\mathbf{W}_{s} \in \R^{(L \cdot D) \times H}$;
  Our Patch-based Parallel Decoding \method{}~\ding{184} decodes each patch to the time domain using a small shared patch-level linear layer $\mathbf{W}_{p} \in \R^{D \times P}$ without modeling temporal relationships among patches. 
  Specifically, the parameters of our patch-based decoding layer are only $\frac{P}{L * H}$ (\emph{e.g.}, $0.34\%$, when $P=16, L=64, H=720$), compared to the sequence-based decoding layer.
  \textbf{Learning Paradigms.} Instead of contrastive learning, masking reconstruction, and limited fine-tuning of the LLMs~\ding{172}, we adopt a forecasting-based two-stage pre-training task~\ding{173} to better transfer the sequence modeling capabilities within LLMs to time series.
  }
  \vspace{-5mm}
  \label{fig-decoder}
\end{figure*}

The advent of large language models (LLMs) has revolutionized the field of natural language processing \citep{chatgpt,llm}. Their remarkable adaptability extends beyond text, as demonstrated through prompting or fine-tuning for various modalities \citep{humantomato,Audiolm}. This adaptability has sparked a surge of interest in leveraging LLMs for TSA. Some studies have explored the use of frozen LLMs in TSA, either through the artful design of prompts \cite{zero-time-learner,ltm-finance,PromptCast} or by reprogramming input time-series \citep{Time-llm,TEMPO}. Others have experimented with fine-tuning LLMs for specific TSA tasks \cite{onefitsall,Llm4ts,TEST}. While these methods show promise, they tend to fall short in generating a comprehensive time-series representation due to implicit representation adaption and inappropriate sequence-wise decoder~\citep{ltm-first,independent_patch} in \shortautoref{fig-decoder}.

\iffalse
Recently, large language models (LLMs) have demonstrated remarkable capabilities in natural language processing~\citep{chatgpt,llm}. 
Moreover, through prompting or fine-tuning, LLMs can be adeptly adapted to modeling sequence modalities other than texts~\citep{humantomato,Audiolm}. %, positioning them as a potential framework for time-series analysis. 
Consequently, there has been a growing interest in applying LLMs for time-series analysis. One group of works uses frozen LLMs for TSA by carefully designing prompts~\cite{zero-time-learner,ltm-finance,PromptCast} or reprogramming the input time-series~\citep{Time-llm,TEMPO}. 
Other works finetune the LLM backbone for specific time-series tasks~\cite{onefitsall,Llm4ts,TEST}. 
While delivering promising results, only performing prompt optimization or finetuning LLMs for specific TSA tasks cannot generate a general time series representation. 
\fi

Recognizing the potential of LLMs in time-series modeling and the shortcomings of the aforementioned methods, we present \emph{\method{}}, an innovative framework that fully realizes the potential of LLMs for general time-series representation learning. Our framework reconceptualizes time-series forecasting as a self-supervised, multi-patch\footnote{The patch concept, introduced by \citet{PatchTST}, denotes the segmentation of the original time series at the subseries level, serving as input tokens for transformer-based TSA models.} prediction task. This approach offers a more effective mechanism for capturing temporal dynamics at the patch level, mitigates the modeling inconsistency in contrastive learning and the temporal dependencies disruption in mask-and-reconstruction, and aligns with the casual pre-training process of LLMs.
% standing in contrast to traditional time-series pre-training methodologies such as mask-and-reconstruction or contrastive learning.

Specifically, we implement a two-pronged self-supervised training strategy to tailor LLMs for time-series representation learning. The first stage involves causal continual training on a variety of time-series datasets, focusing on next-patch prediction to synchronize LLM capabilities with time-series data intricacies. The subsequent stage involves fine-tuning the model for multi-patch prediction in targeted time-series scenarios. A pivotal aspect of our framework is the innovative design of the patch-wise decoder (depicted in Figure~\ref{fig-decoder}). This design mandates the model to decode each patch independently into temporal sequences, deviating from the conventional sequence-wise decoding approach, thus enabling the encoding of time-series representations directly within patches as the decoder is precluded from using the patch sequence for temporal dynamics modeling.

\iffalse

For the pre-training task, we design a two-stage forecasting-based pre-training strategy that adapts the LLMs to the time-series modality. 
% 
Initially, we employ a causal continual pre-training based on next-patch prediction to transfer LLMs' sequence representation capabilities from natural language to the new time-series modality. Subsequently, we reframe time series forecasting as a self-supervised, multi-patch prediction task, optimizing LLM's sequence representation in the target time-series context by leveraging look-back window patches as history prompts.
% 
Regarding the representation decoder, We disentangle the encoding and decoding in the patch-based time series modeling with a patch-wise decoder, empowering the LLM backbone and patch-wise decoder to excel in their designated roles: optimizing patch representation and decoding each patch independently.
\fi

% 
In summary, the primary contributions of this work include:
\vspace{-5mm}
\begin{itemize}
    \vspace{-2mm}
    \item We introduce \emph{\method{}}, an innovative framework adapting LLMs for patch-based time-series representation learning. This framework utilizes a two-stage forecasting-based pre-training strategy. The first stage encompasses causal next-patch training, transferring LLM capabilities for nuanced understandings of time-series data, followed by a fine-tuning stage focused on multi-patch prediction, ensuring a robust representation adaptation to specific time-series contexts.
    \vspace{-1mm}
    \item Diverging from traditional approaches that utilize sequence-wise decoding in TSA tasks, we propose a novel patch-wise decoding methodology. This approach significantly improves the adaptability of LLM backbones, optimizing patch-based time-series representation learning more effectively.
    % \vspace{-1mm}
    \item \emph{\method{}} demonstrates superior performance across various downstream TSA tasks and diverse time-series data domains, validating its ability to derive time-series representations with remarkable transferability and setting new benchmarks in the field.
\end{itemize}
% \vspace{-3mm}

%% file: sections/2-related_work.tex
\section{RELATED WORK}\label{sec-related_Work}

\subsection{Time Series Representation Learning}
The field of time series representation learning has witnessed increasing interest in recent years, with self-supervised learning methods playing a pivotal role. These methods generally fall into two categories:

\textbf{Contrastive Learning.}
This category encompasses methods designed to refine the representation space by leveraging various forms of consistency, such as subseries consistency~\citep{representation,som-vae}, temporal consistency~\cite{tnc,CoST,ts2vec}, transformation consistency~\citep{transformation-cl-1,btsf}, and contextual consistency~\citep{ts-tcc}. The goal is to ensure that representations of positive pairs are closely aligned, whereas those of negative pairs are distinctly separated. %A notable example, TS2Vec~\citep{ts2vec}, segments time series into patches to apply contrastive loss both at the instance and patch levels. 
Despite their strengths, contrastive learning methods often struggle with aligning to low-level tasks such as forecasting, primarily due to their focus on high-level information~\citep{bad-contrasive-learning}. 

\textbf{Masked Modeling.}
%Masked modeling is predicated on the concept of learning abstract representations through the reconstruction of masked periods from their unmasked counterparts. 
PatchTST~\citep{PatchTST} pioneers the utilization of patches as the basic unit for processing time series, advocating for the prediction of masked subseries-level patches to grasp local semantic information while minimizing memory consumption. %Similarly, SimMTM~\citep{SimMTM} aims to reconstruct the entire time series from segments that have been randomly masked. 
However, this approach tends to compromise temporal dependencies and fails to guarantee adequate representation learning of temporal dynamics, as the masked values are often predictably inferred from adjacent contexts~\citep{Ti-MAE}.

\vspace{-1mm}
\subsection{Time Series Analysis based on LLMs}
\vspace{-0.5mm}
The adaptation of pre-trained LLMs for time series analysis has garnered significant attention, exploiting their exceptional ability in sequence representation learning. This body of work can be categorized into three primary approaches:
\vspace{-10pt}
\begin{itemize}
    \item \textbf{Zero-shot Adaptation.} Studies such as those by \citet{zero-time-learner,PromptCast,ltm-finance} leverage the inherent sequence representation capabilities of frozen LLMs to facilitate time series prediction without any task-specific training.
    \vspace{-5pt}
    \item \textbf{Prompt Optimization.} Works by \citet{Time-llm,TEMPO,TEST} employ reprogrammed input time series in conjunction with the innate sequence modeling prowess of pre-trained LLMs, aiming for enhanced forecasting accuracy.
    \vspace{-5pt}
    \item \textbf{Limited Fine-tuning.} Research efforts like those of \citet{onefitsall,ltm-health,Llm4ts} apply fine-tuning to selected components of LLMs to improve performance in time series analysis tasks. 
\end{itemize}
\vspace{-7pt}
While these approaches yield encouraging outcomes, they predominantly focus on distinct TSA tasks, instead of achieving a holistic time-series representation.

\iffalse

\subsection{Time Series Analysis based on LLMs}
Recently, many works have been trying to adapt pre-trained LLMs' excellent sequence representation learning ability in time series analysis. 
(1) \textbf{Zero-shot Adaption.}
\citet{zero-time-learner,PromptCast,ltm-finance} directly adapt the sequence representation ability of frozen LLMs for obtaining time series prediction.
(2) \textbf{Prompt Optimization.}
\citet{Time-llm,TEMPO,TEST} combine the reprogrammed input time series with the inherited sequence modeling ability within pre-trained LLMs to achieve future forecasting.
(3) \textbf{Limited Finetuning.}
\citet{onefitsall,ltm-health,Llm4ts} fine-tune specific parts of LLMs in target datasets for better time series analysis task performance.
% 
However, these methods end with relatively poor performance due to limited adapting strategies~\citep{TEST,ltm-first}, where a representation adaption pre-training and an appropriate patch-wise decoder are needed to transfer the sequential representation capabilities of LLM into time series analysis.
\fi

%% file: sections/3-preliminaries_and_motivation.tex
\vspace{-1mm}
\section{PRELIMINARIES AND MOTIVATION}\label{sec-preliminaries}
\vspace{-0.5mm}
\subsection{Preliminaries}
\vspace{-0.5mm}
\textbf{Time-Series Forecasting~(TSF)}
\label{sec:tsf}
is the fundamental challenge in time series analysis~\cite{ma2023survey}, aiming to analyze the
dynamics and correlations among historical time-series data to predict future
behavior, formulated as:
% \vspace{-2mm}
\begin{equation}
    \resizebox{.70\hsize}{!}
    {
        \label{eq-ts_model}
        \begin{math}
        P(\mathbf{x}_{i}^{L+1:L+H} | \mathbf{x}_{i}^{1:L})
        =
        \prod_{t=L+1}^{L+H} P(x_{i}^{t} | \mathbf{x}_{i}^{1:t-1})
        \end{math}
    }
\end{equation}
where $L$ is the look-back window size and $x_{i}^{t}$ is the value of the $i_{th}$ variate at the $t_{th}$ time step, and the modeling target is to learn the unknown distribution of the $H$ future values.

\textbf{Casual Language Model Pre-training.}
\label{sec:casual_modeling}
Current LLMs mostly belong to casual language models~\citep{chatgpt,gpt2}. They utilize a diagonal masking matrix, ensuring that each token can only access information from previous tokens. The training objective is to predict the next token based on the history information, defined as:
\vspace{-0.5mm}
\begin{equation}
\label{Eq:casual-loss}
\resizebox{.60\hsize}{!}
{$\mathcal{L}_{CLM} = 
\sum_{i=2}^{N} \text{log} P(\mathbf{x}_{i} | \mathbf{x}_{1}, \cdots, \mathbf{x}_{i-1})
$}
\end{equation}
where $N$ is the number of tokens, $\mathbf{x}_i$ denotes the $i$-th token.

\vspace{-1mm}
\subsection{Motivation}
\vspace{-0.5mm}
We explain the motivation of our proposed \method{} solution by raising and answering the following two questions.

\begin{figure*}[t]
  \centering
  \includegraphics[width=.98\textwidth]{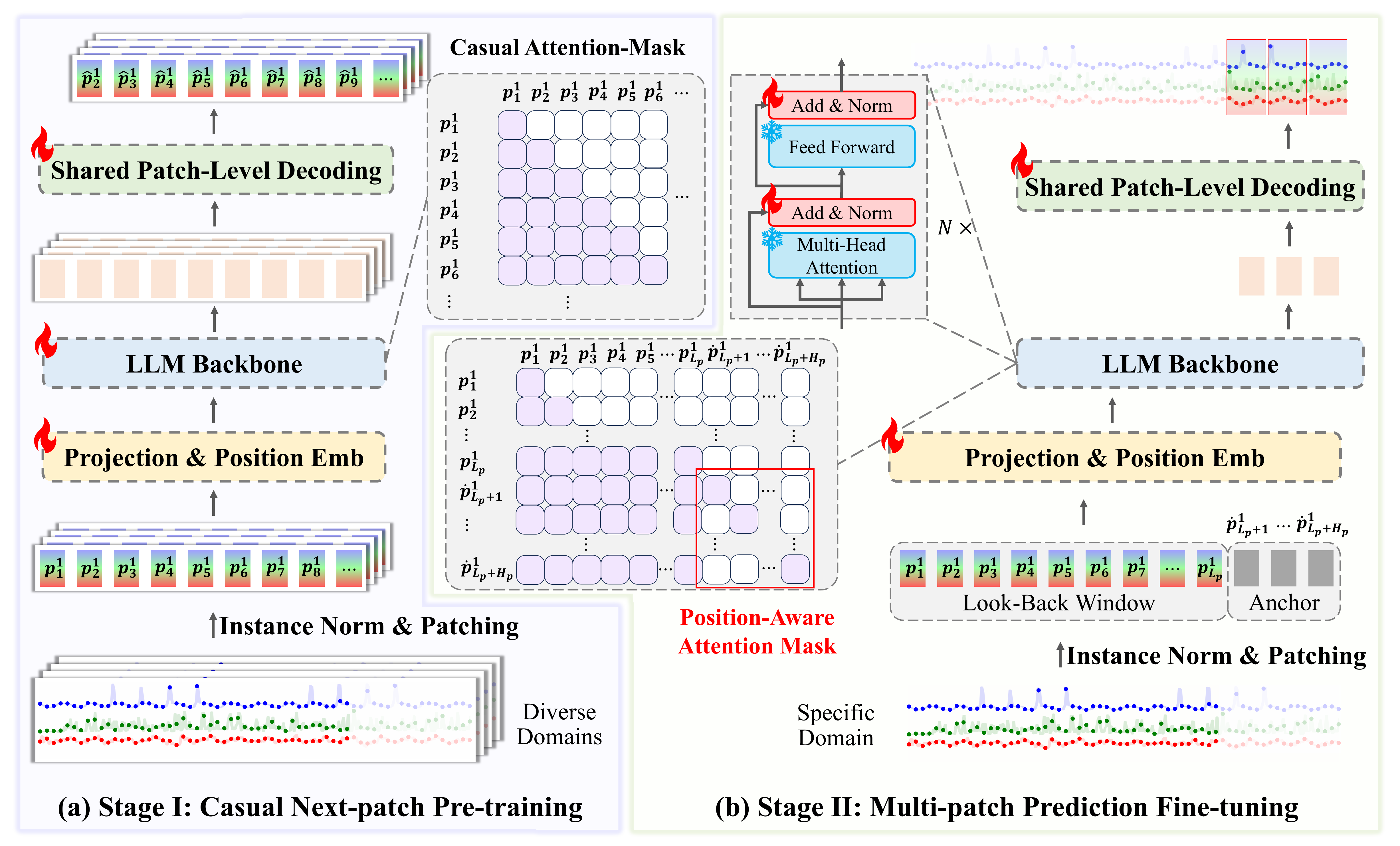}
  \vspace{-3mm}
  \caption{
  \textbf{The model framework of \method{}}. 
  % stage 1: Casual continual pre-training 
  In stage 1, \textbf{Casual Next-patch Pre-training~(a)}, time series from different datasets are initially converted into univariate patch sequences. Then, we conduct next-patch prediction training with casual attention, effectively syncing LLM capabilities with the intricacies of time-series data.
  % stage 2: Multi-patch prediction fine-tuning
  In stage 2, \textbf{Multi-patch Prediction Fine-tuning~(b)}, we fine-tune a few layers for multi-patch prediction in the target time-series context. Firstly, non-parametric methods are first employed to obtain the initial anchor representation of the horizon. Next, we concatenate the look-back window patches and anchors and feed them into the time-series-aligned LLM after stage 1 training with a position-aware attention mask, optimizing anchors with history patches. Finally, all optimized horizon anchors are independently decoded into the target temporal domain through a shared patch-wise linear decoder.
  }
  \vspace{-4mm}
  \label{fig-method}
\end{figure*}

\textbf{How can we effectively adapt LLMs in time series modality?}
\label{motivation-1} 
% 
% \vspace{-3mm}
Traditional contrastive learning~$\mathcal{L}_{CL}$ and mask-and-reconstruction~$\mathcal{L}_{MR}$ serve as a training loss, defined as:
\begin{equation}
    \resizebox{.9\hsize}{!}
    {
    \begin{math}
        \begin{aligned}
        \mathcal{L}_{CL} &= 
        - \frac{1}{N} \sum_{i=1}^{N} \text{log} \frac{\text{exp}(f(\mathbf{x}_i)^{\text{T}}f(\mathbf{x}_i^p))}{\text{exp}(f(\mathbf{x}_i)^{\text{T}}f(\mathbf{x}_i^p)) + \sum_{j=1}^{B-1} \text{exp}(f(\mathbf{x}_i)^{\text{T}}f(\mathbf{x}_i^j))}, \\
        \mathcal{L}_{MR} &= 
        \sum_{i=1}^{N} 
        \|\mathbf{x}_{i}-\widehat{\mathbf{x}}_{i}\|_{2}^{2}=\sum_{i=1}^{N} 
        \|\mathbf{x}_{i}-f(\text{Mask}(\mathbf{x}_{i}))\|_{2}^{2},
    \end{aligned}
    \end{math}
    }
\end{equation}
where $N$ denotes the number of training samples~(pairs), $B-1$ is the number of negative samples, $\mathbf{x}_i$ denotes the $i$-th sample, $\mathbf{x}_i^p$ is the only positive sample, $\mathbf{x}_i^j$ is the $j$-th negative sample of $\mathbf{x}_i$, $\widehat{\mathbf{x}}_{i}$ is the corresponding masked sample, $f(\cdot)$ is the forward pass.
Both are inappropriate for adapting LLMs into time series since the non-negligible misalignment between their modeling objective with time series modeling and casual sequence pre-training processes of LLMs~\citep{bad-contrasive-learning,SimMTM}.
However, during the pre-training stages in \shortautoref{Eq:casual-loss}, casual LLMs undergo a similar training process as TSF in \shortautoref{eq-ts_model} on massive tokens~\citep{llm,gpt2}, where $x_{i}^{t}$ is the $t_{th}$ token of the $i_{th}$ sentences.
This drives us to reformulate time-series forecasting as a self-supervised multi-patch prediction task. 
This offers several benefits:
(1)~\textbf{Guarantee modeling consistency} between high-level representation optimization and downstream low-level tasks, where contrastive learning fails, fostering a versatile representation with robust predictive capabilities excelling in diverse TSA tasks.
(2)~\textbf{Avoid the temporal dependencies disruption} caused by random masking in masked modeling.
(3)~\textbf{Align with LLMs' pre-training} where each token is casually predicted, facilitating the seamless adaptation of LLM to the temporal domain.
Thus, we devise a forecasting-based self-supervised training strategy, as in \shortautoref{fig-method}~(a) and (b), to naturally sync LLMs' excellent representation learning capabilities with time-series variations, including a casual next-patch continual pre-training and a fine-tuning for multi-patch prediction in the target time-series context.

\vspace{-1mm}

\textbf{Is the sequence-level decoder suitable for patch-based time series representation?}
Current TSA models based on LLMs~\citep{Time-llm,onefitsall} follow the traditional patch-based framework in \shortautoref{fig-decoder}~\ding{183}. 
Given patch-based time series representation $\{\vp_1, \cdots, \vp_{L_p} \}, \vp_i \in \R^D$ across $L_p$ patches with dimension $D$, they concatenate and map the patch sequence to prediction horizon $H_p$, through a sequence-level decoder $\mathbf{W}_{s} \in \R^{(L_p \cdot D) \times H_p}$, which can be particularly oversized if either one or all of these values are large, causing severe downstream task overfitting.
% 
% And this also leads to the model's inability to support arbitrary lengths of input~($L_p$)~/output~($H_p$).
% 
Instead of mapping at the sequence level, we disentangle the encoding and decoding within our framework through a patch-wise decoder in \shortautoref{fig-decoder}~\ding{184}, which is involved throughout our pipeline~(Stage 1 and Stage 2). This empowers the LLM backbone and patch-wise decoder to excel in its designated role: encoding each patch for better representation and decoding each patch independently to the temporal domain.

%% file: sections/4-method.tex
\vspace{-5mm}
\section{METHOD}\label{sec-method}
\vspace{-1mm}

Our \method{} stands as a novel framework in redefining the landscape of adapting LLMs into time series analysis. 

\vspace{-2mm}
\subsection{Casual Next-patch Continual Pre-Training\label{sec:stage1}}
\vspace{-0.5mm}
In this section, we propose to conduct casual next-patch continual pre-training, to sync pre-trained LLMs sequence modeling capabilities with time-series modalities on diverse time series datasets(e.g., Weather, Traffic), as in \shortautoref{fig-method}~(a). 

\textbf{Forward Process.} Given time series from various datasets, we first flatten them into $M$ univariate sequences. We denote the $i$-th univariate series of look-back window size $L$ starting at time index $t$ as $\vx^{(i)}_{t:t+L-1} = \{x_{t}^{(i)},..., x_{t+L-1}^{(i)}\}\in \R^{1 \times L}$ where $i=1,...,M$. 
Then each of them is first divided into patch sequence $\vp^{(i)}_{t_{p}:t_{p}+L_{p}-1} = \{ \vp_{t_{p}}^{(i)},..., \vp_{t_{p}+L_{p}-1}^{(i)}\} \in \R^{L_{p} \times P}$ where $t_{p} = \lfloor \frac{(t-P)}{S} \rfloor + 1$ is starting patch index, $L_{p} = \lfloor \frac{(L-P)}{S} \rfloor + 1$ is the number of patches, $P$ is the patch length, and $S$ is the sliding stride.
Finally, each sequence is fed independently into the casual LLM backbone, such as GPT2~\citep{gpt2} for the channel-independence setting. Then we get the casual next-patch prediction $\hat{\vp}^{(i)}_{t_{p}+1:t_{p}+L_p}= \{\hat{\vp}^{(i)}_{t_{p}+1},..., \hat{\vp}^{(i)}_{t_{p}+L_p}\} \in \R^{L_p \times D}$.

\textbf{Loss Function.} We choose to use the MSE loss to guide the representation alignment at the patch level. The loss in each time series is gathered and averaged over $M$ time series to get the overall objective loss: 
$
\mathcal{L}_{p} = \E_{\vp} \frac{1}{M}\sum_{i=1}^M \| \hat{\vp}^{(i)}_{t_{p}+1:t_{p}+L_{p}} - \vp^{(i)}_{t_{p}+1:t_{p}+L_{p}} \|_2^2
$
, which is employed to guide the casual next-patch time-series representation optimization to uncover the hidden temporal dynamics while aligning with LLMs sequential modeling abilities.

\subsection{Multi-patch Prediction Fine-tuning\label{sec:stage2}}
\vspace{-0.5mm}
In this section, we fine-tune for the self-supervised, multi-patch prediction task, further refining patch representations to align with the target time series temporal contexts based on the casual next-patch pre-training model in \shortautoref{sec:stage1}.

\textbf{Forward Process.} 
As illustrated in \shortautoref{fig-method} (b), given the $i$-th univariate time series $\vx^{(i)}_{t:t+L-1} = \{x_t^{(i)},..., x_{t+L-1}^{(i)}\}\in \R^{1 \times L}$ of look-back window size $L$ starting at time index $t$, the prediction horizon $H$ and the time-series-aligned LLM $f_{\theta}(\cdot)$ trained in \shortautoref{sec:stage1}, we firstly prepare prediction anchors $\dot{\vx}^{(i)}_{t+L:t+L+H-1} = \{\dot{x}_{t+L}^{(i)},..., \dot{x}_{t+L+H-1}^{(i)}\}\in \R^{1 \times H}$ through non-parametric methods~(Recent history $\vx^{(i)}_{t+L-H:t+L-1}$or Discrete fourier prediction).
Then the look-back window input and anchors are divided into patch sequence $\vp^{(i)}_{t_{p}:t_{p}+L_{p}-1} = \{ \vp_{t_{p}}^{(i)},..., \vp_{t_{p}+L_{p}-1}^{(i)}\} \in \R^{L_{p} \times P}$ and $\dot{\vp}^{(i)}_{t_{p}+L_{p}:t_{p}+L_{p}+H_{p}-1} = \{ \dot{\vp}_{t_{p} + L_{p}}^{(i)},..., \dot{\vp}_{t_{p}+L_{p}+H_{p}-1}^{(i)}\} \in \R^{H_{p} \times P}$ where $t_{p} = \lfloor \frac{(t-P)}{S} \rfloor + 1$, $L_{p} = \lfloor \frac{(L-P)}{S} \rfloor + 1$, $H_{p} = \lfloor \frac{(H-P)}{S} \rfloor + 1$, $P$ is the patch length, and $S$ is the sliding stride.
Next, these two patch sequences are concatenated and fed into the time-series-aligned LLM backbone $f_{\theta}(\cdot)$ trained in stage 1 with a position-aware attention mask $\mathbf{A}_{p}$, which enhances the temporal relationships among patches (Each future anchor can only see all accurate history patches and itself.). 
Finally, we employ the patch-wise projection layer~$\mathbf{W}_{p} \in \R^{D \times P}$ to independently decode optimized anchors $\dot{\vp}^{(i)_{o}}_{t_{p}+L_{p}:t_{p}+L_{p}+H_{p}-1} \in \R^{H_{p} \times D}$ into temporal patches~$\hat{\vp}^{(i)}_{t_{p}+L_{p}:t_{p}+L_{p}+H_{p}-1}\in \R^{H_{p} \times P}$, formulated as $\hat{\vp}_{t_{p} + L_{p}+k}^{(i)} = \dot{\vp}_{t_{p} + L_{p}+k}^{(i)_{o}}\mathbf{W}_{p}, k \in [0, H_{p}-1]$.

\textbf{Loss Function.} 
We flatten the predicted patches from $\hat{\vp}^{(i)}_{t_{p}+L_{p}:t_{p}+L_{p}+H_{p}-1}$ to $\hat{\vx}^{(i)}_{t+L:t+L+H-1}$, gather and average the loss in $M$ time series: 
$
\mathcal{L}_{s} = \E_{\vx} \frac{1}{M}\sum_{i=1}^M \| \hat{\vx}^{(i)}_{t+L:t+L+H-1} - \vx^{(i)}_{t+L:t+L+H-1} \|_2^2.
$
Notably, during multi-patch~(anchor) representation optimization, most parameters in the time-series-aligned LLM $f_{\theta}(\cdot)$ trained in \shortautoref{sec:stage1} are frozen except for the Position Embedding and Layer Normalization Layer~(Less than $0.01\%$ of the overall parameters) to make better adaptions in target time series. 
Also, once finish a single stage 2 adaption in a target time series dataset, history~$L$, horizon~$H$, we can perform any other forecasting tasks with other input/out length $\hat{L}/\hat{H}$ without any re-training since our patch-wise representation decoding is independent of input/out length.

%% file: sections/5-experiments.tex
\vspace{-4mm}
\section{EXPERIMENTS}\label{sec-experiments}
\vspace{-1mm}
\method{} consistently outperforms state-of-the-art time series analysis methods (Sec.~\ref{sec:Experimental_Settings}) across multiple benchmarks and task settings, including long-term and short-term forecasting (Sec.~\ref{sec:Long-Term_Time_Series_Forecasting} and Sec.~\ref{sec:Short-Term_Time_Series_Forecasting}), few-shot forecasting (Sec.~\ref{sec:Few-shot_Time_Series_Forecasting}), and anomaly detection (Sec.~\ref{sec:Time_Series_Anomaly_Detection}).  
We compared \method{} against a broad collection of models, including the state-of-the-art LLM-based time series analysis model GPT4TS~\citep{onefitsall}.\footnote{To ensure fair comparisons, all experimental configurations are the same as \citet{onefitsall} and follow a unified evaluation pipeline: \url{https://github.com/thuml/Time-Series-Library}. 
For current LLM-based methods~(GPT4TS~\citep{onefitsall} and Time-LLM~\citep{Time-llm}) that are not included in the pipeline, we reproduced their results through their official publicly available code~(\url{https://github.com/DAMO-DI-ML/NeurIPS2023-One-Fits-All} and \url{https://github.com/KimMeen/Time-LLM}). Notably, we use GPT-2~\citep{gpt2} as the default backbone for all LLM-based methods.}
Notably, in forecasting, based on our patch-wise decoder, \method{} excels at handling arbitrary look-back window sizes and prediction horizons \textbf{with only one uniform training setting}, whereas previous methods necessitate re-training for each setting.
Then, we provide additional analysis of representation learning ability and ablation study in Sec.~\ref{sec:Representation_Learning} and Sec.~\ref{sec:Ablation_Study}. Due to the page limit, more experiment results are in the appendix, including imputation, classification, and other exploration experiments. 
We use the same default LLM backbone GPT-2 with the first $6$ layers as GPT4TS~\citep{onefitsall}.

% GPT-2~\citep{gpt2} with the first $6$ layers serves as the default LLM backbone unless otherwise specified.
\input{tables/long-term-forecasting}

\input{tables/short-term-forecasting}
\vspace{-2mm}
\subsection{Experimental Settings}
\vspace{-1mm}
\label{sec:Experimental_Settings}

\textbf{Datasets.} For long-term forecasting, few-shot forecasting, and representation learning, we evaluate our proposed \method{} on $8$ popular datasets, including Weather, Traffic, Electricity, ILI, and $4$ ETT datasets (ETTh1, ETTh2, ETTm1, ETTm2). 
These datasets have been extensively used and are publicly available in ~\citet{autoformer}. 
For short-term forecasting, we evaluate models on widely used marketing dataset M4~\citep{m4}.
For anomaly detection, we evaluate models on five widely employed datasets: SMD~\cite{SMD}, MSL~\cite{MSL_SMAP}, SMAP~\cite{MSL_SMAP}, SWaT~\cite{SWaT}, and PSM~\cite{PSM}. 

\textbf{Baselines.} For time series forecasting and anomaly detection task, we compare $11$ baseline methods: the SOTA LLM-based model GPT4TS~\cite{onefitsall}, nine Transformer-based models, including PatchTST~\cite{PatchTST}, FEDformer~\citep{fedformer}, Autoformer~\citep{autoformer}, Non-Stationary Transformer~\cite{station}, ESTformer~\cite{estformer}, LightTS~\citep{lightTS}, Pyraformer~\citep{pyraformer}, Reformer~\citep{reformer}, Informer~\citep{Informer}, and the SOTA MLP-based model DLinear~\citep{DLinear}. 
And we also add a strong text-guided LLM baseline Time-LLM~\citep{Time-llm} for long-term forecasting.
Besides, N-HiTS~\cite{nhits} and N-BEATS~\cite{nbeats} are added for comprehensive short-term forecasting performance comparison.
For representation learning, we compare \method{} with $5$ baseline methods: the SOTA masking-based representation learning method PatchTST~\citep{PatchTST}, and four contrastive learning methods for time series, BTSF \citep{btsf}, TS2Vec \citep{ts2vec}, TNC \citep{tnc}, and TS-TCC \citep{ts-tcc}.

\textbf{Experiment Settings.} For the time series forecasting task, all models follow the same experimental setup with prediction length $H\in \{24, 36, 48, 60\}$ for the ILI dataset and $H\in \{96, 192, 336, 720\}$ for other datasets. We use the default look-back window $L=336$ for all baseline models and our proposed framework \method{}.

\textbf{Metrics.} We use the following metrics in the experiment comparison: Mean Square Error (MSE), Mean Absolute Error (MAE), Symmetric Mean Absolute Percentage Error (SMAPE), and F1-score~\citep{grishman1996message}.

\input{tables/tab_few_shot_main.tex}
\input{tables/anomaly}

\vspace{-2mm}
\subsection{Long-Term Time Series Forecasting}
\label{sec:Long-Term_Time_Series_Forecasting}

As shown in the table \ref{tab:long-term-forecasting}, \method{} outperforms all baseline methods in most cases. 
Specifically, our informative two-stage forecasting-based, self-supervised representation optimization for LLMs and patch-wise decoding lead to an average performance improvement of $\mathbf{9.71\%}$ over GPT4TS, the current SOTA LLM-based method which directly employs the sequence modeling capabilities in LLMs without any representation adaption.
Compared to Time-LLM, which combines textual prompts with the sequence modeling capabilities of frozen LLMs, \method{} achieves over $\mathbf{9.40\%}$ performance improvement.
Compared with the SOTA Transformer model PatchTST, \method{} realizes an average MSE reduction of $2.03\%$. 
Our improvements are also noteworthy compared with the other model classes, e.g., DLinear or TimesNet, exceeding $\mathbf{19.3\%}$.
Notably, in stage 1, we conduct a shared casual next-patch pre-training with training sets of Weather, Traffic, Electricity, ILI, and 4 ETT datasets.
In stage 2, for $4$ horizons $H\in \{96, 192, 336, 720\}$ in a target dataset, \method{} performs \textbf{only one} training for $H=720$, then it can forecast arbitrary horizons due to its patch-wise decoder that is independent with horizon, while other baselines have to fine-tune for each horizon length.

\vspace{-2mm}
\subsection{Short-Term Time Series Forecasting}
\label{sec:Short-Term_Time_Series_Forecasting}

Table \ref{tab:short-term-forecasting} shows the results of short-term forecasting in the M4 benchmark, which contains marketing data in different frequencies. 
Specifically, we use the same backbone as \shortautoref{sec:Long-Term_Time_Series_Forecasting} which undergoes a casual next-patch pre-training on training sets of Weather, Traffic, Electricity, ILI, and 4 ETT datasets. Then we employ the same model configuration as GPT4TS to fine-tune for each frequency.
We achieve a competitive performance close to the current SOTA TimesNet, whose CNN-based structure is usually considered to perform better in datasets characterized by diverse variations but limited volume~\citep{timesnet}, such as M4.
The overall $\textbf{0.472}$ SMAPE reduction compared with SOTA LLM-based GPT4TS is attributed to the importance of syncing LLM capabilities with the temporal dynamics. This also verifies the excellent transferability of our forecasting-aligned representation across time series from different domains.

\vspace{-0.2cm}
\subsection{Few-shot Time Series Forecasting}
\label{sec:Few-shot_Time_Series_Forecasting}
\vspace{-0.1cm}

LLMs have demonstrated remarkable performance in few-shot learning~\cite{ltm-health,llm} due to their ability to obtain strong general representations. 
In this section, we evaluate the few-shot forecasting ability of our time-series-aligned LLM in ETT datasets.
To avoid data leakage, we conduct stage 1 on ETT1 datasets and perform multi-patch prediction with only $\textbf{5}\%$ training data in ETT2, and vice versa.
The few-shot forecasting results are shown in \shortautoref{tab:few_shot_main}. 
\method{} remarkably excels over all baseline methods, and we attribute this to the successful representation syncing in our two-stage representation adaption. Notably, both our \method{} and GPT4TS consistently outperform other competitive baselines, further verifying the potential of LLMs as effective time series machines.
Significantly, \method{} achieves an average MSE reduction of $\textbf{8.6}\%$ compared to SOTA LLM-based GPT4TS, indicating the benefits of our forecasting-based adaption and patch-wise decoding. 
In comparison to convolution-based TimesNet and MLP-based DLinear models that are usually considered more data-efficient for training and suitable for few-shot learning methods, \method{} still demonstrates an average MSE reduction of $\textbf{30.6}\%$ and $\textbf{28.5}\%$ respectively.

\vspace{-0.2cm}
\subsection{Time Series Anomaly Detection}
\label{sec:Time_Series_Anomaly_Detection}
\vspace{-0.05cm}

Time series anomaly detection has various industrial applications, such as health monitoring or finance evaluation.
Similar to short-term forecasting, we use the same backbone as \shortautoref{sec:Long-Term_Time_Series_Forecasting} which undergoes a casual next-patch pre-training on training sets of Weather, Traffic, Electricity, ILI, and 4 ETT datasets. Then we employ the same model configuration as GPT4TS to fine-tune in each anomaly dataset.
Results in \shortautoref{tab:anomaly} demonstrate that \method{} achieves the SOTA performance with the averaged F1-score $\textbf{87.51}\%$. 
We attribute this better capability of detecting infrequent anomalies within time series to the forecasting-aligned representation learned in stage 1 casual next-patch pre-training on diverse time series.
The aligning process syncs the LLM sequence modeling abilities with time series and further enhances the representation's transferability and generality across various time series domains and downstream tasks.

\vspace{-0.2cm}
\subsection{Representation Learning}
\label{sec:Representation_Learning}
\vspace{-0.05cm}

\input{tables/representation_main}

In addition to \method{}'s outstanding performance across various downstream tasks, we further explore its superiority in adapting LLMs for time-series representation learning in ETTh1 forecasting. This is achieved through comparisons with state-of-the-art representation learning methods, and various comparative experiments for our two-stage forecasting-based \method{}.
Detailed results are in \shortautoref{table::compare with self-sup}. 

\textbf{${\text{Sta}_{1}}$: Casual Next-patch Pre-training.} 
Comparing the column ${\text{Sta}_{1}}$+${\text{Sta}_{2}}$ with column Masking+${\text{Sta}_{2}}$\footnote{Explanation of column names can be found in table caption.}, we observe a distinct average performance decline over $\textbf{13.5}\%$, whether in a pre-trained LLM or a vanilla PatchTST. We attribute it to the challenge that masking-based patch pre-training struggles to model the vital temporal variations due to its random masking and reconstruction training paradigm.
The results of other contrastive-learning-based methods also lead to a significant performance decline, as a result of the non-negligible non-alignment between their high-level objective with the downstream time series analysis tasks.

\textbf{${\text{Sta}_{2}}$: Multi-patch Prediction Fine-tuning.}
Previous patch-based models~\citep{PatchTST,onefitsall} all choose to concatenate and project in sequence-level as in \shortautoref{fig-decoder}~\ding{183}. 
Comparing column Masking+SD and Masking+${\text{Sta}_{2}}$, more than $\textbf{13.4}\%$ deterioration occurs, strongly indicating the great risks of overfitting in the huge sequence-wise decoder.

\vspace{-0.1cm}

\input{tables/ablation-study}
\subsection{Ablation Study} 
\label{sec:Ablation_Study}

\vspace{-0.05cm}

In this section, we conduct several ablations on framework design and the effectiveness of our two-stage forecasting-based self-supervised training. 
Brief results are in \shortautoref{tab:ablations}.

\textbf{Casual Next-Patch Continual Pre-training.}
Comparing row A.1 and B.1 in \shortautoref{tab:ablations}, an average MSE increase of $\textbf{8.80}\%$ is observed, indicating that ablating casual next-patch continual pre-training significantly harms the sequence pattern recognition and forecasting modeling of the LLM for effective time series analysis. 
We attribute it to the inadequate adaption to apply pre-trained LLMs in time series without alignment that fits the temporal dynamics, forecasting modeling, and the casual pre-training of LLMs.

\textbf{LLM Pre-trained Weight.}
We designed two sets of ablation experiments with different model sizes to avoid the mismatch between training data and model parameter quantity.
We discard the pre-trained weights of the LLMs and train from scratch the first $6$ layers~(\textbf{B.2}) and the first $3$ layers~(\textbf{B.3}) of GPT-2. 
Ablating the LLM pre-trained weights directly results in the loss of the learned sequential representation capabilities from massive sequential text data~\citep{onefitsall,zero-time-learner}. 
Consequently, it becomes difficult to learn the temporal representation from scratch within the LLM architecture, leading to the degradation in performance of $\textbf{5.15}\%$ and $\textbf{7.91}\%$, respectively.

\textbf{Patch-level Decoder.}
In ablation experiment \textbf{C.1}, we employed the conventional sequence-level decoder, resulting in an average performance loss exceeding $\textbf{8.54}\%$. 
Despite using a decoder over $100$ times larger and can train specifically for each input/output length, a substantial performance loss occurred. 
This is attributed to the potential downstream task overfitting of the huge sequence-level head and the incapability to disentangle the patch representation encoding and decoding process, leading to inadequate patch representation optimization in the LLM backbone.

\textbf{Position-aware Attention Mask.}
In \method{}, we transform the forecasting into multi-patch representation optimization based on well-aligned patch-based time series knowledge. Position-aware attention mask is designed to further enhance the optimization process by removing the unwanted confusion brought by other being-optimized anchors during the optimization. Ablation of this component (\textbf{C.2}) results in over $\textbf{10.01}\%$ performance deterioration.

\subsection{Interpretability Experiment\label{sec:Interpretability-Experiment}}

\begin{figure}[!htbp]
\centering
\vspace{-5mm}
\includegraphics[width=\columnwidth]{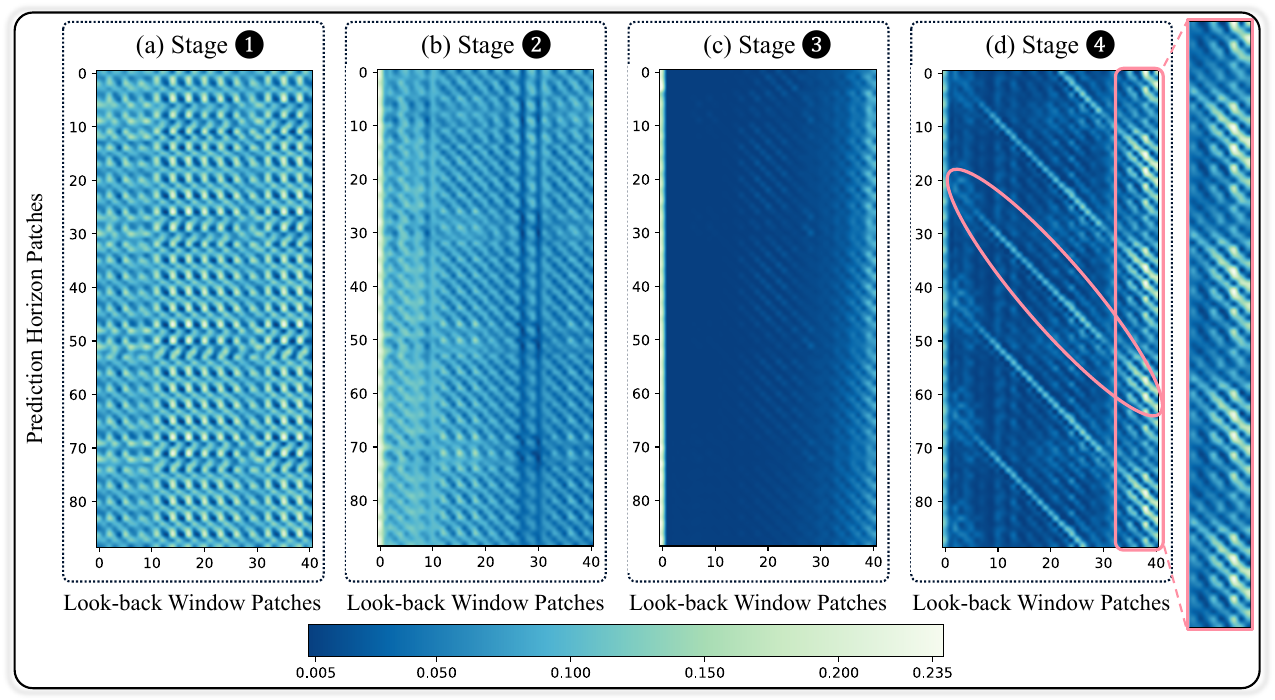}
\vspace{-8mm}
\caption{Interpretability study in Traffic dataset. Due to the page limit, we put the full visualization and analysis in the appendix. The Y-axis and X-axis represent prediction horizon patch indexes and look-back window patch indexes, respectively.}
\vspace{-3mm}
\label{fig:Interpretability}
% \vspace{-20pt}
\end{figure}

We conducted a case study on the Traffic dataset to illustrate the evolution of attention weights from the prediction horizon patches to look-back window patches at four stages in \shortautoref{fig:Interpretability}. 
The 4 subplots detail the attention weights optimization process from randomly-initialized~(Stage~\ding{182}), through LLMs-pretrained~(Stage~\ding{183}), casually next-patch continual pre-trained~(Stage~\ding{184}) to multi-patch prediction adaption~(Stage~\ding{185}).
Our observations are as follows:
\textbf{Obs.\ding{172} After stage \ding{185}, \method{} adeptly captures the complex multi-periodic properties of time series and a discernible trend of increasing information importance along the temporal dimension.}
In \shortautoref{fig:Interpretability}~(d), look-back window patches closest to the prediction horizon exhibit similar patterns from prediction horizon patches at time steps $t$, $t+3$, $\cdots$. With a patch size of 16 and a stride of 8, sampling hourly, this corresponds to local day cycles. Additionally, there exist 20-patch cycles (equivalent to 168 hours), indicating weekly cycles. Furthermore, look-back window patches closer to the predicted horizon receive increasing attention due to their temporal proximity, indicating their greater informational significance.
\textbf{Obs.\ding{173} After stage \ding{184}, \method{} learns universal single-period features (e.g., day) and showcases a noticeable trend of increasing attention along the time dimension} in \shortautoref{fig:Interpretability}~(c), stemming from the process of casually predicting the next patch.
\textbf{Obs.\ding{174} Pre-trained LLM parameters capture fundamental time-series cycle attributes} in \shortautoref{fig:Interpretability}~(a) and (b), serving as a robust optimization anchor for time-series representation learning when compared with random initialization.

%% file: tables/long-term-forecasting.tex
\begin{table*}[ht]
\vspace{-4mm}
\caption{\textbf{Long-term Forecasting Results.} We calculate the MSE for each dataset. A lower value indicates better performance. {\boldres{Red}}: the best, \secondres{Underlined}: the second best. Due to the page limit, we put the full table in the appendix.}
\label{tab:long-term-forecasting}
% \vspace{1mm}
\centering
\small
\scalebox{0.79}{
\setlength\tabcolsep{2.5pt}
\begin{tabular}{c|c|c|c|c|c|c|c|c|c|c|c|c|c|c}
\toprule
\multicolumn{2}{c|}{\textbf{Methods}} & \textbf{\method{}} & \textbf{GPT4TS} & \textbf{Time-LLM} & \textbf{DLinear} & \textbf{PatchTST} & \textbf{TimesNet} & \textbf{FEDformer} & \textbf{Autoformer} & \textbf{Stationary} & \textbf{ETSformer} & \textbf{LightTS} & \textbf{Informer} & \textbf{Reformer} \\
\midrule
\multirow{4}{*}{\rotatebox{90}{ETTh1}}
& 96  &0.380 & \secondres{0.376} & 0.399 & \boldres{0.375} &\boldres{0.375} &0.384&0.376&0.449&0.513&0.494&0.424&0.865&0.837\\
& 192 &\boldres{0.396}  & 0.416 & 0.433 & \secondres{0.405}  & 0.414 &0.436&0.420&0.500&0.534&0.538&0.475&1.008&0.923\\
& 336 &\boldres{0.413} & 0.442 & 0.469 &0.439& \secondres{0.431} &0.491&0.459&0.521&0.588&0.574&0.518&1.107&1.097\\
& 720 & \secondres{0.461}  & 0.477 & 0.473 & 0.472  & \boldres{0.449} &0.521&0.506&0.514&0.643&0.562&0.547&1.181&1.257\\
\midrule
\multirow{4}{*}{\rotatebox{90}{ETTh2}}
& 96  &\boldres{0.251}  & 0.285 & 0.294 & 0.289  & \secondres{0.274}&0.340&0.358&0.346&0.476&0.340&0.397&3.755&2.626\\
& 192 &\boldres{0.298}  & 0.354 & 0.355 & 0.383 & \secondres{0.339} &0.402&0.429&0.456&0.512&0.430&0.520&5.602&11.12\\
& 336 &\secondres{0.343} & 0.373 & 0.372 & 0.448  & \boldres{0.331} &0.452&0.496&0.482&0.552&0.485&0.626&4.721&9.323 \\
& 720 & 0.417  & \secondres{0.406} & 0.428 & 0.605  & \boldres{0.379} &0.462&0.463&0.515&0.562&0.500&0.863&3.647&3.874\\
\midrule
\multirow{4}{*}{\rotatebox{90}{ILI}}
& 24  & \boldres{1.359}  & 2.063 & 1.617 & 2.215  & \secondres{1.522} &2.317& 3.228 & 
3.483 &2.294 & 2.527 & 8.313 & 5.764 & 4.400 \\
& 36 & \boldres{1.405}  & 1.868 & 1.708 & 1.963  & \secondres{1.430} & 1.972 & 2.679 &3.103 &1.825 &2.615 &6.631 &4.755 & 4.783\\
& 48 & \boldres{1.442}  & 1.790 & 1.633 & 2.130  &  \secondres{1.673} & 2.238 & 2.622 & 2.669 &2.010 &2.359 &7.299 &4.763&4.832\\
& 60 & \secondres{1.603}  & 1.979 & 2.106 & 2.368  & \boldres{1.529} &2.027&2.857&2.770&2.178&2.487&7.283&5.264&4.882 \\
\midrule
\multirow{4}{*}{\rotatebox{90}{Weather}}
& 96  & \boldres{0.149}   & 0.162 & 0.163 & 0.176 & \secondres{0.152} & 0.172 & 0.217 & 0.266 & 0.173 & 0.197 &  0.182 & 0.300 & 0.689 \\
& 192 &\boldres{0.190}   & 0.204 & 0.206 & 0.220 & \secondres{0.197}  & 0.219 & 0.276 & 0.307 & 0.245 & 0.237 & 0.227 & 0.598 & 0.752 \\
& 336 &\boldres{0.238}   & 0.254 & 0.255 & 0.265 & \secondres{0.249} & 0.280 & 0.339 & 0.359 & 0.321 & 0.298 & 0.282 & 0.578 & 0.639 \\
& 720 & \boldres{0.316}   & 0.326 & 0.325 & 0.333 & \secondres{0.320} & 0.365 & 0.403 & 0.419 & 0.414 & 0.352 & 0.352 & 1.059 & 1.130 \\
\midrule
\multirow{4}{*}{\rotatebox{90}{Traffic}}
& 96  & \secondres{0.372}   & 0.388 & 0.383 & 0.410 & \boldres{0.367} &  0.593 & 0.587 & 0.613 & 0.612 & 0.607 & 0.615 & 0.719 & 0.732 \\
& 192 &\boldres{0.383}   &  0.407 & 0.398 &  0.423 & \secondres{0.385} & 0.617 & 0.604 & 0.616 & 0.613 & 0.621 & 0.601 & 0.696 & 0.733 \\
& 336 &\boldres{0.396}   & 0.412 & 0.407 & 0.436 & \secondres{0.398} & 0.629 & 0.621 & 0.622 & 0.618 & 0.622 & 0.613 & 0.777 & 0.742 \\
& 720 & \boldres{0.433}   & 0.450 & \secondres{0.434} & 0.466 & \secondres{0.434} & 0.640  & 0.626 & 0.660 & 0.653 & 0.632 & 0.658 & 0.864 & 0.755 \\
\midrule
\multirow{4}{*}{\rotatebox{90}{Electricity}}
& 96  & \boldres{0.127}   & 0.139 & 0.140 & 0.140 & \secondres{0.130} & 0.168 & 0.193 & 0.201 & 0.169 & 0.187 & 0.207 & 0.274 & 0.312\\
& 192 &\boldres{0.145}   & 0.153 & 0.151 & 0.153 & \secondres{0.148} & 0.184 & 0.201 & 0.222 & 0.182 & 0.199 & 0.213 & 0.296 & 0.348 \\
& 336 &\boldres{0.163}   & 0.169 & 0.171 & 0.169 & \secondres{0.167} & 0.198 & 0.214 & 0.231 & 0.200 & 0.212 & 0.230 & 0.300 & 0.350\\
& 720 & \secondres{0.206}   & 0.206 & 0.210 & 0.203 & \boldres{0.202} & 0.220 & 0.246 & 0.254 & 0.222 & 0.233 & 0.265 & 0.373 & 0.340\\
\bottomrule
\end{tabular}
}
\vspace{-5mm}
\end{table*}

%% file: tables/short-term-forecasting.tex
\begin{table*}[ht]
\renewcommand\arraystretch{1.5}
\caption{\textbf{Short-term Forecasting Results.} We calculate the SMAPE for each dataset. A lower value indicates better performance. {\boldres{Red}}: the best. IMP. denotes the absolute SMAPE reduction of \method{} compared with SOTA LLM-based GPT4TS, where a larger value indicates a better improvement. Due to the page limit, we put the full table in the appendix.}
\label{tab:short-term-forecasting}
%\vskip 0.15in
\vspace{-4.5mm}
\begin{center}
% \begin{small}
\scalebox{0.68}{
\setlength\tabcolsep{3pt}
\begin{tabular}{c|c|cccccccccccccc}
\toprule

Methods&IMP.&\method{}&GPT4TS&TimesNet&PatchTST&N-HiTS&N-BEATS& ETSformer& LightTS& DLinear &FEDformer &Stationary &Autoformer  &Informer&Reformer \\

\midrule
$Yearly$
&\textbf{1.264}& 13.540  &14.804&\boldres{13.387}&13.477&13.418&13.436&18.009&14.247&16.965&13.728&13.717&13.974&14.727&16.169\\
\midrule

$Quarterly$
&\textbf{0.292}& 10.216 &10.508&\boldres{10.100}&10.38&10.202&10.124&13.376&11.364&12.145&10.792&10.958&11.338&11.360&13.313\\
\midrule

$Monthly$
&\textbf{0.206}& 12.775 &12.981&\boldres{12.670}&12.959&12.791&12.677&14.588&14.014&13.514&14.260&13.917&13.958&14.062&20.128\\
\midrule

$Others$
&\textbf{0.250}& 5.032 &5.282&\boldres{4.891}&4.952&5.061&4.925&7.267&15.880&6.709&4.954&6.302&5.485&24.460&32.491\\
\midrule

$Average$
&\textbf{0.472}& 11.950
&12.422 &\boldres{11.829}&12.059& 11.927& 11.851& 14.718& 13.525& 13.639 &12.840 &12.780 &12.909 &14.086 &18.200 \\

\bottomrule

\end{tabular}
}
% \end{small}
\end{center}
\vspace{-8mm}
\end{table*}

%% file: tables/tab_few_shot_main.tex
\begin{table*}[ht]
\vspace{-2mm}
\caption{\textbf{Few-shot Forecasting on $5\%$ Data.} We calculate the MSE and MAE for each dataset. All results are averaged from 4 prediction lengths (96, 192, 336, and 720). \boldres{Red}: the best, \secondres{Underlined}: the second best. Due to the page limit, we put the full table in the appendix.}
\label{tab:few_shot_main}
\centering
\small
\scalebox{0.72}{
\setlength\tabcolsep{3pt}
\begin{tabular}{c|cc|cc|cc|cc|cc|cc|cc|cc|cc|cc|cc|cc}
\toprule

\multirow{2}{*}{Methods} 
&\multicolumn{2}{c|}{\method{}}&\multicolumn{2}{c|}{GPT4TS}  & \multicolumn{2}{c|}{DLinear}&\multicolumn{2}{c|}{PatchTST}&\multicolumn{2}{c|}{TimesNet}&\multicolumn{2}{c|}{FEDformer}&\multicolumn{2}{c|}{Autoformer}&\multicolumn{2}{c|}{Stationary}&\multicolumn{2}{c|}{ETSformer}&\multicolumn{2}{c|}{LightTS}&\multicolumn{2}{c|}{Informer}&\multicolumn{2}{c}{Reformer} \\
&MSE&MAE&MSE&MAE&MSE&MAE&MSE&MAE&MSE&MAE&MSE&MAE&MSE&MAE&MSE&MAE&MSE&MAE&MSE&MAE&MSE&MAE&MSE&MAE \\

\midrule

ETTh1    
& \boldres{0.608} &\boldres{0.507} &0.681&\secondres{0.560}&0.750&0.611&0.694&0.569&0.925&0.647&\secondres{0.658}&0.562&0.722&0.598&0.943&0.646&1.189&0.839&1.451&0.903&1.225&0.817&1.241&0.835\\
ETTh2     
& \boldres{0.374} & \boldres{0.417} &\secondres{0.400}&\secondres{0.433}& 0.694&0.577&0.827&0.615&0.439&0.448&0.463&0.454&0.441&0.457&0.470&0.489&0.809&0.681&3.206&1.268&3.922&1.653&3.527&1.472\\
ETTm1      
& \secondres{0.419}&\boldres{0.414}&0.472&0.450&\boldres{0.400}&\secondres{0.417}&0.526&0.476&0.717&0.561&0.730&0.592&0.796&0.620&0.857&0.598&1.125&0.782&1.123&0.765&1.163&0.791&1.264&0.826\\
ETTm2     
& \boldres{0.297}&\boldres{0.345}&\secondres{0.308}&\secondres{0.346}&0.399&0.426&0.314&0.352&0.344&0.372&0.381&0.404&0.388&0.433&0.341&0.372&0.534&0.547&1.415&0.871&3.658&1.489&3.581&1.487\\
\midrule

Average     
& \boldres{0.425} & \boldres{0.421}& \secondres{0.465} & \secondres{0.447}
& 0.594 & 0.517 & 0.493 & 0.461  & 0.612 &  0.509  & 0.553 &  0.504  & 0.594 &  0.535  & 0.653 &  0.518  & 0.914 &  0.712  & 1.799 &  0.952  & 2.492 &  1.188  & 2.403 &  1.155 \\

\bottomrule

\end{tabular}
}
\vspace{-5mm}
\end{table*}

%% file: tables/anomaly.tex
\begin{table*}[ht]
\caption{\textbf{Anomaly Detection Results.} We calculate the F1-score (as \%) for each dataset. \boldres{Red}: the best,  \secondres{Underlined}: second best. $\ast$. in the Transformers indicates the name of $\ast$former. Due to the page limit, we put the full table in the appendix.}
\vspace{-5mm}
\label{tab:anomaly}
\begin{center}
\begin{small}
\scalebox{0.89}{
\setlength\tabcolsep{3pt}
\begin{tabular}{c|ccccccccccccccc}
\toprule

\multirow{2}{*}{Methods} 
& \method{} 
& \multirow{2}{*}{GPT4TS} & \multirow{2}{*}{TimesNet} & \multirow{2}{*}{PatchTS.} & \multirow{2}{*}{ETS.} & \multirow{2}{*}{FED.} & \multirow{2}{*}{LightTS} & \multirow{2}{*}{DLinear} & \multirow{2}{*}{Stationary} & \multirow{2}{*}{Auto.} & \multirow{2}{*}{Pyra.}& \multirow{2}{*}{In.} & \multirow{2}{*}{Re.} & \multirow{2}{*}{LogTrans.} & \multirow{2}{*}{Trans.} \\
&Ours&&&&&&&&&&&&&& \\

\midrule
SMD
& \secondres{85.42} 
&\boldres{86.89} &84.61 &84.62&83.13& 85.08& 82.53& 77.10 &84.72 &85.11 &83.04&81.65 &75.32& 76.21& 79.56 \\
MSL 
& 82.26 
&82.45&81.84&78.70&\boldres{85.03}&78.57&78.95&\secondres{84.88}&77.50&79.05&84.86&84.06&84.40&79.57&78.68 \\
SMAP
& \boldres{78.04}
&\secondres{72.88}&69.39&68.82& 69.50& 70.76& 69.21& 69.26& 71.09& 71.12& 71.09& 69.92& 70.40& 69.97& 69.70 \\
SWaT
& \boldres{94.57}
&\secondres{94.23}& 93.02&85.72 & 84.91& 93.19 &93.33& 87.52& 79.88& 92.74& 91.78& 81.43& 82.80& 80.52& 80.37 \\
PSM
& 97.19 
&97.13&  \boldres{97.34} &96.08& 91.76& 97.23& 97.15& 93.55& \secondres{97.29} & 93.29& 82.08& 77.10& 73.61 &76.74 &76.07 \\
\midrule
Average 
&\boldres{87.51}
&\secondres{86.72}&85.24&82.79& 82.87& 84.97& 84.23& 82.46& 82.08& 84.26& 82.57& 78.83& 77.31& 76.60& 76.88\\
\bottomrule
\end{tabular}
}
\end{small}
\end{center}
\vspace{-6mm}
\end{table*}

%% file: tables/representation_main.tex
\begin{table*}[!htbp]
\centering
    \vspace{-2mm}
    \caption{
    \textbf{Representation Learning Methods Comparison.} 
    ${\text{Sta}_{1}}$ implies conducting stage 1, casual continual pre-training based on next-patch prediction in diverse time series. 
    ${\text{Sta}_{2}}$ implies conducting stage 2, multi-patch prediction fine-tuning for target time series. 
    % % 
    $\text{Masking}-\ast\%$ implies replacing the origin stage 1 with masking-patch pre-training of masking ratio $\ast\ast\%$~(\textit{Masking} in PatchTST denotes using default masking ratio $40\%$ as its origin paper). 
    \textit{SD} denotes replacing the patch-wise decoder in stage 2 with sequence-wise decoder.
    {\boldres{Red}}: the best, \secondres{Underlined}: the second best.
    IMP. denotes the improvement on best results of \method{} compared to that of baselines.
    }
    \vspace{1mm}
    \scalebox{0.685}{
    \setlength{\tabcolsep}{1.0mm}{
        \begin{tabular}{c|c|c|cccccc|cccccc|cccccccc}
            \toprule
            \multicolumn{2}{c|}{\multirow{2}{*}{Models}}& \multirow{2}{*}{IMP.}& \multicolumn{6}{c|}{\method{}}& \multicolumn{6}{c|}{PatchTST}&  \multicolumn{2}{c|}{\multirow{2}{*}{BTSF}}&\multicolumn{2}{c|}{\multirow{2}{*}{TS2Vec}}&\multicolumn{2}{c|}{\multirow{2}{*}{TNC}}&\multicolumn{2}{c}{\multirow{2}{*}{TS-TCC}}\\
            \cline{4-15}
            \multicolumn{2}{c|}{}& &\multicolumn{2}{c|}{${\text{Sta}_{1}}$+${\text{Sta}_{2}}$}& \multicolumn{2}{c|}{Masking-20\%+${\text{Sta}_{2}}$}&\multicolumn{2}{c|}{Masking-40\%+${\text{Sta}_{2}}$}& \multicolumn{2}{c|}{${\text{Sta}_{1}}$+${\text{Sta}_{2}}$}& \multicolumn{2}{c|}{Masking+SD}& \multicolumn{2}{c|}{Masking+${\text{Sta}_{2}}$}& \multicolumn{2}{c|}{} & \multicolumn{2}{c|}{} & \multicolumn{2}{c|}{} & \\
            \midrule
            \multicolumn{2}{c|}{Metrics}& MSE &MSE&MAE&MSE&MAE&MSE&MAE&MSE&MAE&MSE&MAE&MSE&MAE&MSE&MAE&MSE&MAE&MSE&MAE&MSE&MAE\\
            \midrule
            \multirow{5}*{\rotatebox{90}{ETTh1}} & 24  & 6.52$\%$ &\boldres{0.301} & \boldres{0.362}& 0.417 & 0.435 & 0.441 & 0.447 & 0.402 & 0.433 & \secondres{0.322} & \secondres{0.369} & 0.431 & 0.453 & 0.541 & 0.519 & 0.599 & 0.534 & 0.632 & 0.596 & 0.653 & 0.610 \\
            \multicolumn{1}{c|}{}& 48 & 3.48$\%$ &\boldres{0.343} & \boldres{0.377}& 0.418 & 0.437 & 0.422 & 0.439 & 0.401 & 0.432 & \secondres{0.354} & \secondres{0.385} & 0.437 & 0.467 & 0.613 & 0.524 & 0.629 & 0.555 & 0.705 & 0.688 & 0.720 & 0.693 \\
            \multicolumn{1}{c|}{}& 168 & 6.21$\%$ &\boldres{0.393} & \boldres{0.415}& 0.436 & 0.448  & 0.437 & 0.449 & \secondres{0.416} & 0.441 & 0.419 & \secondres{0.424} & 0.459 & 0.480 & 0.640 & 0.532 & 0.755 & 0.636 & 1.097 & 0.993 & 1.129 & 1.044 \\
            \multicolumn{1}{c|}{}& 336 & $7.20\%$ &\boldres{0.413} & \boldres{0.428}& 0.439 & 0.456  & 0.439  & 0.463 & \secondres{0.440} & 0.486 & 0.445 & \secondres{0.446} & 0.479 & 0.483 & 0.864 & 0.689 & 0.907 & 0.717 & 1.454 & 0.919 & 1.492 & 1.076 \\
            \multicolumn{1}{c|}{}& 720 & $5.34\%$ &\boldres{0.461} & \boldres{0.462}& 0.475 & 0.485  & 0.479 & 0.499 & 0.500 & 0.517 & \secondres{0.487} & \secondres{0.478} & 0.523 & 0.508 & 0.993 & 0.712 & 1.048 & 0.790 & 1.604 & 1.118 & 1.603 & 1.206 \\
            \bottomrule			
        \end{tabular}
    }
    }
\label{table::compare with self-sup}
\vspace{-4mm}
\end{table*}

%% file: tables/ablation-study.tex
\begin{table}[!htbp]
% \captionsetup{font=small} 
    \vspace{-4mm}
    \caption{\textbf{Ablations on ETT Dataset} (average MSE in prediction length $[96, 192, 336, 720]$ reported). \boldres{Red}: the best. 
    Due to the page limit, we put the full table and analysis in the appendix.
    }\label{tab:ablations}
    % \vskip -0.05in
    % \vspace{1mm}
    \centering
    \begin{small} 
    \scalebox{0.80}{
    \renewcommand{\multirowsetup}{\centering}
    \setlength{\tabcolsep}{1.0mm}{
    \begin{tabular}{l|cccc}
    \toprule
    \multirow{2}{*}{Model Variant} & \multicolumn{4}{c}{Long-term Forecasting} \\
    \cmidrule{2-5} 
    &ETTh1 &ETTh2 &ETTm1 &ETTm2\\
    \midrule
    \textbf{A.1} \method{} (\textbf{Default}: 6)\tablefootnote{The $*(x)$ denotes using the first $x$ layers of model $*$.} &\boldres{0.413} &\boldres{0.327} &\boldres{0.332} & 0.294 \\
    \textbf{A.2} \method{} (3) & 0.437 & 0.330 & 0.350 & 0.292 \\
    \textbf{A.3} \method{} (9) & 0.452 & 0.339 & 0.404 & 0.307 \\
    \textbf{A.4} \method{} (12) & 0.580 & 0.334 & 0.403 & 0.303 \\
    \midrule
    \textbf{B.1} w/o Casual Continual Pre-training& 0.460 & 0.350 & 0.373 & 0.307 \\
    \textbf{B.2} w/o LLM Pretrained Weights~(6) & 0.437  & 0.336 & 0.364 & 0.301\\
    \textbf{B.3} w/o LLM Pretrained Weights~(3) & 0.462  & 0.339 & 0.373 & 0.305\\
    \midrule
    \textbf{C.1} w/o Patch-level Decoder & 0.455  & 0.387  & 0.362  & \boldres{0.284} \\
    \textbf{C.2} w/o Position-aware Attention Mask & 0.443  & 0.358  & 0.399  & 0.303  \\
    \midrule
    \textbf{D.1} Init with FFT & 0.416  & 0.355  & 0.375  &  0.301 \\
    \textbf{D.2} Init with Random &  0.447 & 0.351  & 0.365  & 0.309  \\
    \midrule
    \textbf{E.1} LN+PE+Attn & 0.442  & 0.346  & 0.363  & 0.319 \\
    \textbf{E.2} LN+PE+Attn+FFN & 0.465  & 0.348  & 0.358  & 0.302 \\
    
    \bottomrule
    \end{tabular}
    }
    }
    \end{small}
\vspace{-3mm}
\end{table}

%% file: sections/6-conclusion.tex
\vspace{-0.1cm}
\section{CONCLUSION}\label{sec-conclusion}

In this paper, we introduce \emph{\method{}}, a novel framework that adapts Large Language Models (LLMs) for time-series representation learning. 
By reframing time-series forecasting as a self-supervised, multi-patch prediction task, our approach surpasses conventional mask-and-reconstruction techniques, more adeptly capturing the hidden temporal dynamics within patch representations. 
Notably, \emph{\method{}} introduces a patch-wise decoding mechanism, a departure from traditional sequence-wise decoding, allowing for the independent decoding of patches into temporal sequences. This significantly bolsters the LLM backbone's aptitude for temporal patch-based representation learning. 
\emph{\method{}} exhibits superior performance across various downstream tasks, substantiating its efficacy in deriving temporal representations with enhanced transferability, marking a pivotal stride in adapting LLMs for advanced time-series analysis.

%% file: sections/7-appendix.tex
\appendix
\onecolumn

% You can have as much text here as you want. The main body must be at most $8$ pages long.
% For the final version, one more page can be added.
% If you want, you can use an appendix like this one.  

% The $\mathtt{\backslash onecolumn}$ command above can be kept in place if you prefer a one-column appendix, or can be removed if you prefer a two-column appendix.  Apart from this possible change, the style (font size, spacing, margins, page numbering, etc.) should be kept the same as the main body.
\section{Visualization}

\begin{figure}[!htbp]
\begin{center}
\includegraphics[width=\columnwidth]{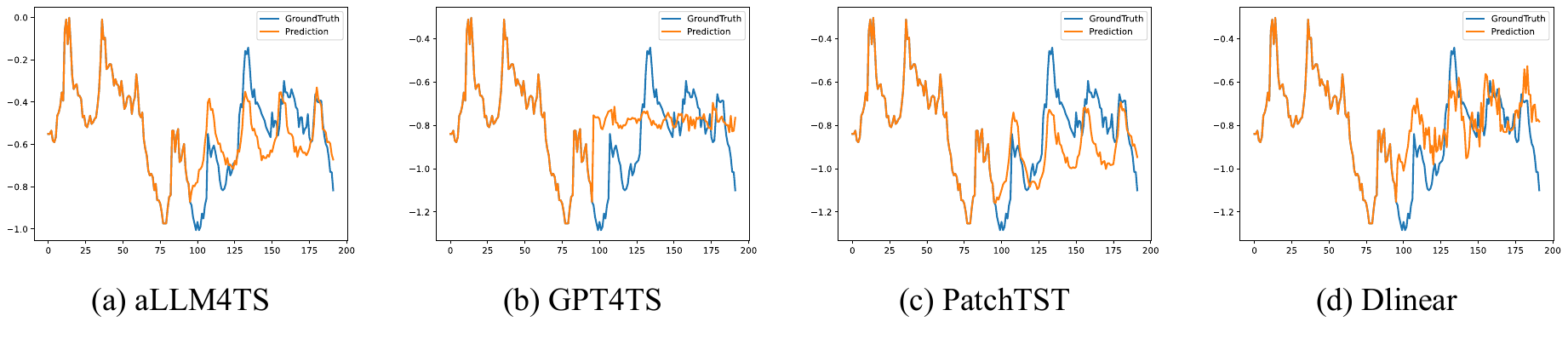}
\vspace{-20pt}
\caption{Long-term forecasting cases from ETTh1 by different models under the input-96-predict-96 settings. \textcolor{blue}{Blue} lines are the ground truths and \textcolor{orange}{orange} lines are the model predictions.}
\label{fig:long-term-forecasting}
\end{center}
% \vspace{-20pt}
\end{figure}

\begin{figure*}[!htbp]
\begin{center}
\includegraphics[width=\columnwidth]{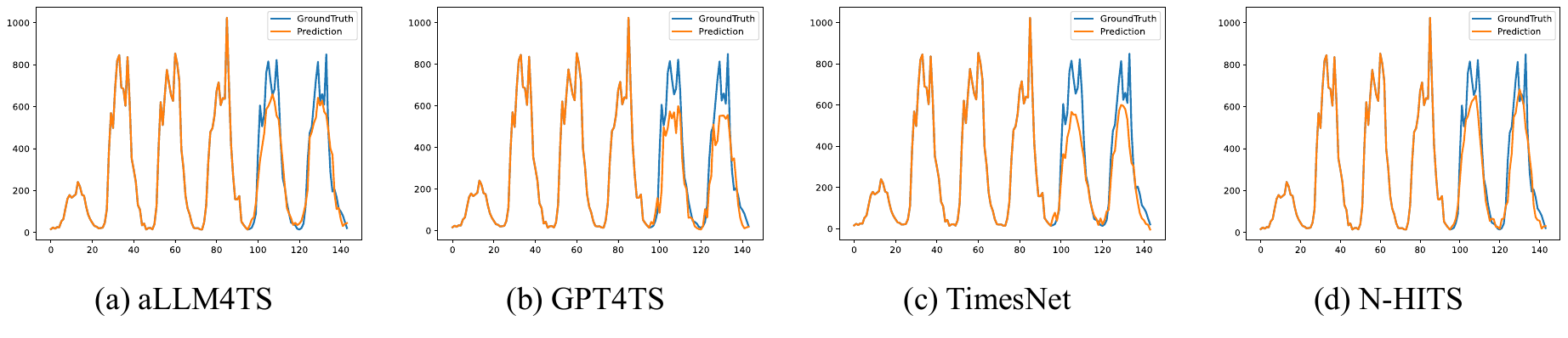}
\vspace{-20pt}
\caption{Short-term forecasting from the M4 hourly dataset by different models under the input-96-predict-48 settings. \textcolor{blue}{Blue} lines are the ground truths and \textcolor{orange}{orange} lines are the model predictions.}
\label{fig:short-term-forecasting}
\end{center}
% \vspace{-20pt}
\end{figure*}

In this section, we present visualizations of the forecasting achieved by \method{} in comparison to state-of-the-art and representative methods, including GPT4TS~\citep{onefitsall}, PatchTST~\citep{PatchTST}, DLinear~\citep{DLinear}, N-HITS~\citep{nhits}, and TimesNet~\citep{timesnet}, for long-term and short-term forecasting. The accompanying figures (\shortautoref{fig:long-term-forecasting} and \shortautoref{fig:short-term-forecasting}) illustrate the comparison between the long-term (input-96-predict-96) and short-term (input-96-predict-48) forecasts of various approaches against the ground truth. Notably, \method{} exhibits superior forecasting accuracy when compared with the state-of-the-art LLM-based GPT4TS~\citep{onefitsall}, transformer-based PatchTST~\citep{PatchTST}, MLP-based DLinear~\citep{DLinear}, MLP-based N-HITS~\citep{nhits} and CNN-based TimesNet~\citep{timesnet}.

\section{More Related Work}

\subsection{Time Series Forecasting}
Time series forecasting has undergone extensive exploration and study as a pivotal task.
From early statistical methods~\cite{ARIMA,Prophet} to recent deep learning approaches based on MLP~\cite{DLinear,Less,nbeats,nhits}, TCN\cite{tcn} or RNN\cite{lstm3,lstm2,lstm}, researchers have been continuously modeling time series forecasting from various perspectives.
% transformers
Recently, transformers~\citep{PatchTST, autoformer, pyraformer, fedformer} have demonstrated remarkable effectiveness in time series forecasting. Leveraging attention mechanisms, they excel in uncovering temporal dependencies among different time points and mining sequence representation. 
% PatchTST
Notably, PatchTST~\citep{PatchTST} first proposes using patches as the fundamental processing unit for time series and explores the capabilities of patch-based representation learning within an self-supervised masking setting.

\subsection{Time Series Representation Learning}
Recently, there has been a growing focus on time series representation learning. Previous self-supervised learning methods can be roughly divided into two classes:
(1) \textbf{Contrastive Learning.}
These methods aim to optimize the representation space based on subseries consistency~\citep{representation,som-vae}, temporal consistency~\cite{tnc,CoST,ts2vec}, transformation consistency~\citep{transformation-cl-1,btsf} or contextual consistency~\citep{ts-tcc}, where representations of positive pairs are optimized to be close to each other and negative ones to be far apart.
For instance, TS2Vec~\citep{ts2vec} splits multiple time series into patches and further defines the contrastive loss in both instance-wise and patch-wise aspects.
TS-TCC~\citep{ts-tcc} optimizes the representation by making the augmentations predict each other's future.
However, these methods suffer from poor alignment with low-level tasks, such as forecasting, due to the main focus on high-level information~\citep{bad-contrasive-learning}.
(2) \textbf{Masked Modeling.}
The fundamental insight behind masked modeling is to learn abstract representation by reconstructing the masked period from the unmasked content~\citep{PatchTST,SimMTM,Ti-MAE}.
PatchTST~\citep{PatchTST}  first proposes using patches as the fundamental processing unit for time series and explores predict masked subseries-level patches to capture the local semantic information and reduce memory usage.
SimMTM~\citep{SimMTM} reconstructs the original time series from multiple randomly masked series.
However, masking modeling not only disrupts temporal dependencies but also fails to ensure sufficient learning of temporal dynamics, as masked values are often easily reconstructed from the surrounding contexts~\citep{Ti-MAE}.

\subsection{Time Series Analysis based on LLMs}
Recently, many works have been trying to adapt pre-trained LLMs' excellent sequence representation learning ability in time series analysis. 
(1) \textbf{Zero-shot Adaption.}
\citet{zero-time-learner,PromptCast,ltm-finance} directly adapt the sequence representation ability of frozen LLMs for obtaining time series prediction.
(2) \textbf{Prompt Optimization.}
\citet{Time-llm,TEMPO,TEST} combine the reprogrammed input time series with the inherited sequence modeling ability within pre-trained LLMs to achieve future forecasting.
(3) \textbf{Limited Finetuning.}
\citet{onefitsall,ltm-health,Llm4ts} fine-tune specific parts of LLMs in target datasets for better time series analysis task performance.
However, these methods end with relatively poor performance due to limited adapting strategies~\citep{TEST,ltm-first}, where a representation adaption pre-training and an appropriate patch-wise decoder are needed to transfer the sequential representation capabilities of LLM into time series analysis.

In this paper, we aim to fundamentally adapt LLMs for time series based on a two-stage forecasting-based self-supervised training strategy, including the casual next-patch pre-training and multi-patch prediction fine-tuning.
Additionally, we devise a patch-wise decoding layer to disentangle the encoding and decoding process in patch-based time series modeling, boosting the LLM backbone to effectively focus on patch-based representation optimization.
Our \method{} stands as a novel framework in redefining the landscape of adapting LLMs into time series analysis.

\section{Dataset Details}

For forecasting and imputation, we utilize eight widely recognized multivariate datasets, as presented in \citet{autoformer}, and their specifics are outlined in \shortautoref{tab:forecasting-data}. 
The \textit{Weather}\footnote{https://www.bgc-jena.mpg.de/wetter/} dataset encompasses 21 meteorological indicators in Germany, while the \textit{Traffic}\footnote{https://pems.dot.ca.gov/} dataset records road occupancy rates from various sensors on San Francisco freeways. 
The \textit{Electricity}\footnote{https://archive.ics.uci.edu/ml/datasets/ElectricityLoadDiagrams20112014} dataset comprises hourly electricity consumption data for 321 customers. 
The \textit{ILI}\footnote{https://gis.cdc.gov/grasp/fluview/fluportaldashboard.html} dataset captures the count of patients and the influenza-like illness ratio weekly. 
The \textit{ETT}\footnote{https://github.com/zhouhaoyi/ETDataset} (Electricity Transformer Temperature) datasets are sourced from two distinct electric transformers labeled 1 and 2, each featuring two resolutions (15 minutes and 1 hour) denoted as "m" and "h". 
Consequently, we have a total of four ETT datasets: \textit{ETTm1}, \textit{ETTm2}, \textit{ETTh1}, and \textit{ETTh2}.

\vspace{-0.5cm}
\begin{table}[!h]
    \centering
    \caption{Statistics of Forecasting and Imputation Benchmark Datasets.}
    \scalebox{1.0}{
    \begin{tabular}{c|cccccccc}
    \toprule  
    Datasets & Weather  & Traffic & Electricity & ILI & ETTh1 & ETTh2 & ETTm1 & ETTm2 \\
    \midrule  
    Features & 21 & 862 & 321 & 7 & 7 & 7 & 7 & 7 \\
    Timesteps & 52696 & 17544 & 26304 & 966 & 17420 & 17420 & 69680 & 69680 \\
    Frequency & 10 min & 15 min & Hourly & Weekly & Hourly & Hourly & 15 min & 15 min \\
    \bottomrule 
    \end{tabular}}
    \label{tab:forecasting-data}
\end{table}

For anomaly detection, we evaluate models on five widely employed datasets: SMD~\cite{SMD}, MSL~\cite{MSL_SMAP}, SMAP~\cite{MSL_SMAP}, SWaT~\cite{SWaT}, and PSM~\cite{PSM}. 
\begin{enumerate}
    \item The Server Machine Dataset (SMD, \cite{SMD}) spans five weeks and originates from a prominent Internet company, encompassing 38 dimensions.
    \item Pooled Server Metrics (PSM, \cite{PSM}) are internally collected from various application server nodes at eBay, encompassing 26 dimensions.
    \item Both the MSL (Mars Science Laboratory rover) and SMAP (Soil Moisture Active Passive satellite) datasets, publicly available from NASA~\citep{MSL_SMAP}, consist of 55 and 25 dimensions, respectively. These datasets encompass telemetry anomaly information extracted from the Incident Surprise Anomaly (ISA) reports of spacecraft monitoring systems.
    \item SWaT (Secure Water Treatment, \cite{SWaT}), is derived from the data collected by 51 sensors within the critical infrastructure system operating continuously.
\end{enumerate}
\begin{table}[h]
  \caption{Statistics of Anomaly Detection Benchmark Datasets~\citep{xu2021anomaly}. Abnormal Proportion represents the true abnormal proportion of the whole dataset.}
  \label{tab:ad-dataset}
  % \vspace{-10pt}
  % \vskip 0.05in
  % \setlength{\tabcolsep}{4.2pt}
  \centering
  \begin{small}
  \begin{tabular}{c|c|cc|ccc|c}
    \toprule
    \scalebox{0.9}{Benchmarks} & \scalebox{0.9}{Applications} & \scalebox{0.9}{Dimension} & \scalebox{0.9}{Window} & \scalebox{0.9}{\#Training} & \scalebox{0.9}{\#Validation} & \scalebox{0.9}{\#Test (labeled)} & \scalebox{0.9}{Abnormal Proportion}  \\
    \midrule
     \scalebox{0.9}{SMD} & \scalebox{0.9}{Server} & \scalebox{0.9}{38} & \scalebox{0.9}{100} & \scalebox{0.9}{566,724} & \scalebox{0.9}{141,681} & \scalebox{0.9}{708,420} & \scalebox{0.9}{0.042}\\
     \scalebox{0.9}{PSM} & \scalebox{0.9}{Server} & \scalebox{0.9}{25} & \scalebox{0.9}{100} & \scalebox{0.9}{105,984} & \scalebox{0.9}{26,497} & \scalebox{0.9}{87,841} &  \scalebox{0.9}{0.278}\\
     \scalebox{0.9}{MSL} & \scalebox{0.9}{Space} & \scalebox{0.9}{55}& \scalebox{0.9}{100} & \scalebox{0.9}{46,653} & \scalebox{0.9}{11,664} & \scalebox{0.9}{73,729} & \scalebox{0.9}{0.105} \\
     \scalebox{0.9}{SMAP} & \scalebox{0.9}{Space} & \scalebox{0.9}{25} & \scalebox{0.9}{100} & \scalebox{0.9}{108,146} & \scalebox{0.9}{27,037} & \scalebox{0.9}{427,617} & \scalebox{0.9}{0.128}\\
     \scalebox{0.9}{SWaT} & \scalebox{0.9}{Water} & \scalebox{0.9}{51} & \scalebox{0.9}{100} & \scalebox{0.9}{396,000} & \scalebox{0.9}{99,000} & \scalebox{0.9}{449,919} & \scalebox{0.9}{0.121} \\
    \bottomrule
  \end{tabular}
  \end{small}
\end{table}

\section{Experimental Details}
\subsection{Implementation}
We adopt the experimental setups introduced by \citet{onefitsall} for all baseline models, ensuring a consistent evaluation pipeline accessible at \url{https://github.com/thuml/Time-Series-Library} to facilitate fair comparisons. GPT-2~\citep{gpt2}, specifically its first $6$ layers, is employed as the default backbone model unless explicitly specified. Our model is implemented in PyTorch~\citep{Pytorch}, and all experiments are executed on NVIDIA A800-80G GPUs. Moreover, we enhance memory efficiency by transforming multivariate data into univariate data, treating each feature of the sequence as an individual time series. This aligns with the demonstrated efficacy of channel independence in prior works such as DLinear~\citep{DLinear} and PatchTST~\citep{PatchTST}.

\subsection{Evaluation Metrics}
We utilize mean square error (MSE) and mean absolute error (MAE) for evaluating long-term forecasting, few-shot forecasting, and imputation. For short-term forecasting on the M4 benchmark, we employ the symmetric mean absolute percentage error (SMAPE), mean absolute scaled error (MASE), and overall weighted average (OWA) following the approach in N-BEATS \citep{nbeats}, where OWA is a metric specific to the M4 competition. Anomaly detection employs Precision, Recall, and F1-score, with the metric calculations outlined as follows:
\begin{align} \label{equ:metrics}
    \text{MSE} &= \frac{1}{H}\sum_{h=1}^T (\mathbf{Y}_{h} - \Hat{\mathbf{Y}}_{h})^2,
    &
    \text{MAE} &= \frac{1}{H}\sum_{h=1}^H|\mathbf{Y}_{h} - \Hat{\mathbf{Y}}_{h}|,\\
    \text{SMAPE} &= \frac{200}{H} \sum_{h=1}^H \frac{|\mathbf{Y}_{h} - \Hat{\mathbf{Y}}_{h}|}{|\mathbf{Y}_{h}| + |\Hat{\mathbf{Y}}_{h}|},
    &
    \text{MAPE} &= \frac{100}{H} \sum_{h=1}^H \frac{|\mathbf{Y}_{h} - \Hat{\mathbf{Y}}_{h}|}{|\mathbf{Y}_{h}|}, \\
    \text{MASE} &= \frac{1}{H} \sum_{h=1}^H \frac{|\mathbf{Y}_{h} - \Hat{\mathbf{Y}}_{h}|}{\frac{1}{H-s}\sum_{j=s+1}^{H}|\mathbf{Y}_j - \mathbf{Y}_{j-s}|},
    &
    \text{OWA} &= \frac{1}{2} \left[ \frac{\text{SMAPE}}{\text{SMAPE}_{\textrm{Naïve2}}}  + \frac{\text{MASE}}{\text{MASE}_{\textrm{Naïve2}}}  \right], \\
    \text{Precision} &= \frac{\Hat{\mathbf{N}}_{TP}}{\Hat{\mathbf{N}}_{TP} + \Hat{\mathbf{N}}_{FP}}
    &
    \text{Recall} &= \frac{\Hat{\mathbf{N}}_{TP}}{\Hat{\mathbf{N}}_{TP} + \Hat{\mathbf{N}}_{FN}}.
\end{align}
$H$ denotes the number of data points, signifying the prediction horizon in our experiments. $s$ denotes the periodicity of the time series data. 
$\mathbf{Y}_{h}$ and $\Hat{\mathbf{Y}}_{h}$ correspond to the $h$-th ground truth and prediction where $h \in \{1, \cdots, H\}$.
$\Hat{\mathbf{N}}_{TP}$ is the number of true positives in prediction, $\Hat{\mathbf{N}}_{FP}$ is the number of false positives in prediction, $\Hat{\mathbf{N}}_{FN}$ is the number of true negatives in prediction, $\text{F1-score} = \frac{2\cdot\text{Precision}\cdot\text{Recall}}{\text{Precision} + \text{Recall}}$.

\subsection{Detailed Definition and Results for Few-shot and Long-term Forecasting}
\label{appendix:few-shot-learning}

Given the validated efficacy of channel independence in time series datasets by \citet{DLinear} and \citet{PatchTST}, we opt for an approach wherein each multivariate series is treated as a distinct independent univariate series. Following standard experimental protocols, we partition each time series into training, validation, and test sets. Specifically, for the few-shot forecasting task, only a designated percentage ($5\%$) of timesteps from the training data are used, while the remaining two parts remain unchanged. The assessment metrics align with those employed in classic multivariate time series forecasting. This experiment is conducted thrice, and the subsequent analyses report the average metrics. Detailed results for few-shot time-series forecasting are presented in Table ~\ref{tab:few-shot-forecasting-5per}.

\input{tables/long-term-forecasting-full}
\clearpage

\input{tables/few-shot-forecasting}

\subsection{Detailed Results for Short-term Forecasting}
\input{tables/short-term-forecasting-full}

Our comprehensive short-term forecasting results are showcased in \shortautoref{tab:short_term_full}. Throughout various scenarios, \method{} consistently outperforms the majority of baseline models. Notably, we achieve a substantial improvement over GPT4TS, with notable margins such as \textbf{3.80\%} overall, \textbf{8.54\%} on M4-Yearly, and an average of \textbf{4.73\%} on M4-Hourly, M4-Daily, and M4-Weekly. In comparison to the recent state-of-the-art forecasting models~(N-HiTS and PatchTST), \method{} also demonstrates comparable or superior performance.

\subsection{Error bars}

All experiments were conducted three times, and we present the standard deviations. Comparisons with the second-best approach, PatchTST~\citep{PatchTST}, for long-term forecasting tasks are elucidated in \shortautoref{tab:errorbar_long}. The table presents the average Mean Squared Error (MSE) and Mean Absolute Error (MAE) for four ETT datasets, accompanied by their corresponding standard deviations.

\input{tables/long-term-forecasitng-with-std}

\subsection{Comparison with Traditional Methods on Few-shot Learning}
\label{app:classical_few_shot}
Deep learning approaches provide benefits over traditional methods for managing extensive datasets; nonetheless, in the context of few-shot learning, it is imperative to also acknowledge the relevance of traditional methods. As depicted in Table \ref{tab:classical_methods}, \method{} still demonstrates superior performance.

\input{tables/zero-classical}

\subsection{Full Abalation Results and Analysis}
\label{app:full_abalation}
\input{tables/abalation-full}
In this section, we conduct several ablations on framework design and the effectiveness of our two-stage forecasting-based pre-training. 
Full results are in \shortautoref{tab:abalation-full}.

\textbf{Casual Next-Patch Continual Pre-training.}
Comparing row A.1 and B.1 in \shortautoref{tab:abalation-full}, an average MSE increase of $\textbf{8.80}\%$ is observed, indicating that ablating casual next-patch continual pre-training significantly harms the sequence pattern recognition and representation modeling of the LLM for effective time series forecasting. 
We attribute it to the inadequate adaption to apply pre-trained LLMs in time series without alignment that fits the time series dynamics and the inherited casual modeling within LLMs.

\textbf{LLM Pre-trained Weight.}
We designed two sets of ablation experiments with different model sizes to avoid the mismatch between training data and model parameter quantity.
We discard the pre-trained weights of the LLMs and train from scratch the first $6$ layers~(\textbf{B.2}) and the first $3$ layers~(\textbf{B.3}) of GPT-2. 
Ablating the LLM pre-trained weights directly results in the loss of the learned sequential representation capabilities from massive sequential text data~\citep{onefitsall,zero-time-learner}. 
Consequently, it becomes difficult to learn the temporal representation from scratch within the LLM architecture, leading to the degradation in performance of $\textbf{5.15}\%$ and $\textbf{7.91}\%$, respectively.

\textbf{Patch-level Decoder.}
In ablation experiment \textbf{C.1}, we employed the conventional sequence-level decoder, resulting in an average performance loss exceeding $\textbf{8.54}\%$. 
Despite using a decoder over $100$ times larger and can train specifically for each input/output length, a substantial performance loss occurred. 
This is attributed to the potential downstream task overfitting of the huge sequence-level head and the incapability to disentangle the patch representation encoding and decoding process, leading to inadequate patch representation optimization in the LLM backbone.

\textbf{Position-aware Attention Mask.}
In \method{}, we transform the forecasting into multi-patch representation optimization based on well-aligned patch-based time series knowledge. Position-aware attention mask is designed to further enhance the optimization process by removing the unwanted confusion brought by other being-optimized anchors during the optimization. Ablation of this component (\textbf{C.2}) results in over $\textbf{10.01}\%$ performance deterioration.

\textbf{The Number of LLM Layers.}
To better balance performance and computational efficiency, we test using various numbers of layers on ETT datasets~(\textbf{A.2}, \textbf{A.3}, \textbf{A.4)}. 
Shallow layers typically operate on modeling coarse-grained token sequence features, which is advantageous for optimizing time series representations~\citep{gpt-layer}. In contrast, deep layers often focus more on domain-specific sequence relationships, which can pose challenges for transferring sequence modeling capabilities to time series.
Thus GPT-2 with 6 layers is chosen as our default backbone.

\textbf{Non-parametric Methods to Initialize the Anchors.}
During stage 2, we employ non-parametric methods to initialize the anchors to be predicted. This initialization includes recent historical values by default and the discrete Fourier decomposition values for the look-back window~(\textbf{D.1}) and random values~(\textbf{D.2}). It is observed that as the noisy information in the initial anchor values increases, the representation optimization becomes more challenging, leading to respective increases in MSE loss of $\textbf{6.16}\%$ and $\textbf{7.65}\%$.

\textbf{Unfrozen Parts in LLM.}
By default, we only unfreeze the layer normalization layer and position embedding to boost downstream forecasting optimization in stage 2, which is considered a common practice~\citep{Frozen_Transformers,Parameter-efficient} to adapt llm in other sequence modeling tasks.
In ablation \textbf{E.1} and \textbf{E.2}, we try to fine-tune other parts like the attention layers or FFN layers but end with the poor performance of more than $\textbf{7.67}\%$ MSE increase. We attribute this to overfitting caused by ineffective training due to limited data in downstream tasks.

\subsection{Full Results of Interpretability Experiment}

\begin{figure*}[!htbp]
\begin{center}
\includegraphics[width=\columnwidth]{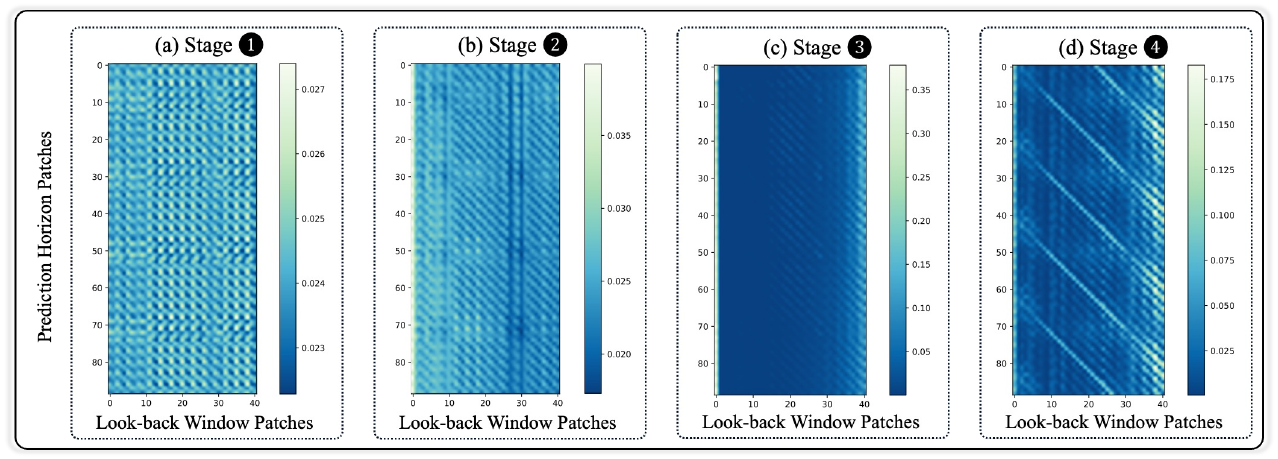}
\vspace{-20pt}
\caption{Case study of Traffic Dataset. The Y-axis and X-axis represent prediction horizon patch indexes and look-back window patch indexes, respectively.}
\label{fig:Interpretability-traffic}
\end{center}
\vspace{-20pt}
\end{figure*}

\begin{figure*}[!htbp]
\begin{center}
\includegraphics[width=\columnwidth]{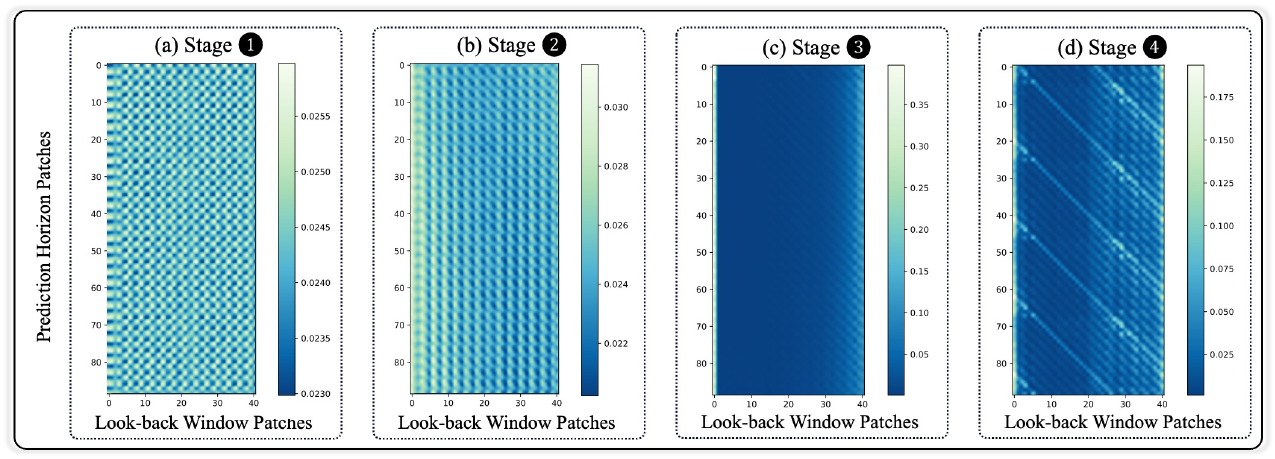}
\vspace{-20pt}
\caption{Case study of Electricity Dataset. The Y-axis and X-axis represent prediction horizon patch indexes and look-back window patch indexes, respectively.}
\label{fig:Interpretability-electricity}
\end{center}
\vspace{-20pt}
\end{figure*}

\begin{figure*}[!htbp]
\begin{center}
\includegraphics[width=\columnwidth]{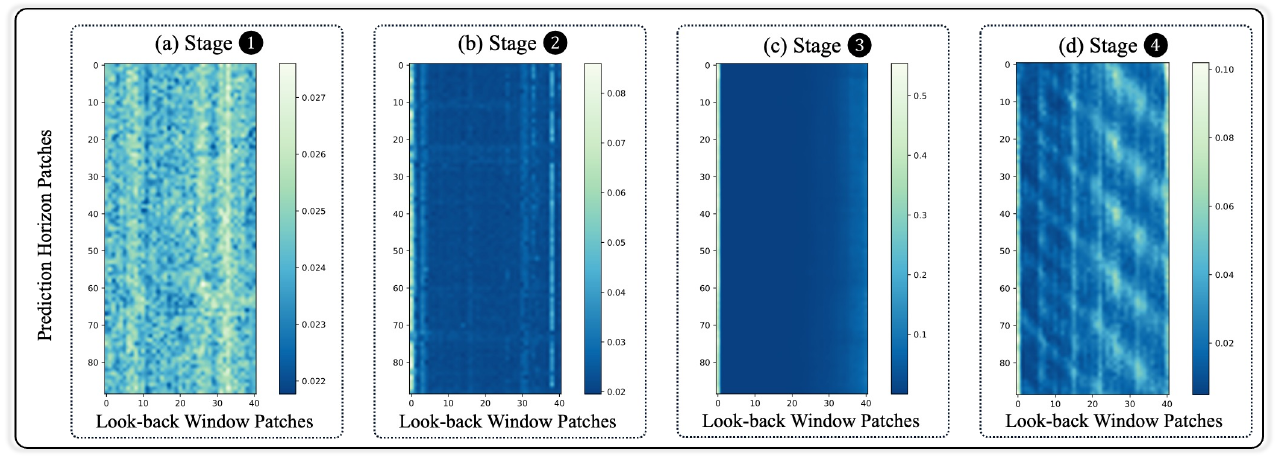}
\vspace{-20pt}
\caption{Case study of Weather Dataset. The Y-axis and X-axis represent prediction horizon patch indexes and look-back window patch indexes, respectively.}
\label{fig:Interpretability-weather}
\end{center}
\vspace{-20pt}
\end{figure*}

We conducted case studies on the Traffic, Electricity, and Weather dataset to illustrate the evolution of attention weights from the prediction horizon patches to look-back window patches at four stages in \shortautoref{fig:Interpretability-traffic}, \shortautoref{fig:Interpretability-electricity} and \shortautoref{fig:Interpretability-weather}. 
The 4 subplots in each figure detail the attention weights optimization process from randomly-initialized~(Stage~\ding{182}), through LLMs-pretrained~(Stage~\ding{183}), casually next-patch continual pre-trained~(Stage~\ding{184}) to multi-patch prediction adaption~(Stage~\ding{185}).
Our whole observations are as follows:
\textbf{Obs.\ding{172} After stage \ding{185}, \method{} adeptly captures the complex multi-periodic properties of time series and a discernible trend of increasing information importance along the temporal dimension.}
This is evidently observed in all \shortautoref{fig:Interpretability-traffic}~(d), \shortautoref{fig:Interpretability-electricity}~(d) and \shortautoref{fig:Interpretability-weather}~(d). 
Among these, look-back window patches closest to the prediction horizon exhibit similar attention patterns from prediction horizon patches at time steps $t$, $t+3$, $\cdots$. With a patch size of 16 and a stride of 8, sampling hourly, this corresponds to local day cycles. Additionally, there exist 20-patch cycles (equivalent to 168 hours), indicating weekly cycles. Furthermore, look-back window patches closer to the predicted horizon receive increasing attention due to their temporal proximity, indicating their greater informational significance.
\textbf{Obs.\ding{173} After stage \ding{184}, \method{} effectively learns universal single-period features (e.g., day) and showcases a noticeable trend of increasing attention along the time dimension}, especially in Traffic in \shortautoref{fig:Interpretability-traffic}~(c) and Electricity in \shortautoref{fig:Interpretability-electricity}~(c) which possess more pronounced periodicity, stemming from the process of casually predicting the next patch.
\textbf{Obs.\ding{174} Pre-trained LLM parameters capture fundamental individual cycle attributes within time series,} offering notable optimization benefits when contrasted with random initialization, thereby serving as a robust optimization reference for downstream time-series representation optimization.

\vspace{-10pt}

\subsection{Further Exploration Experiment on Pre-trained LLMs}

To further demonstrate the significance of pre-training LLM weights for time series representation learning, we highlight that the sequence modeling abilities obtained from extensive text data serve as effective initial anchors for time series representation optimization. In addition to assessing the impact of LLM pre-training parameters in ablation experiments B.2 and B.3 in \shortautoref{sec:Ablation_Study}, and examining the pre-training LLMs' ability to capture fundamental periodicities in time series through the interpretability experiment in \shortautoref{sec:Interpretability-Experiment}, we adopted SOTA transformer-based patchTST~\citep{PatchTST} as the model backbone and conducted same two-stage experiments as \method{} in \shortautoref{sec-method}. The results underscore the utility of pre-training LLM weights as a robust optimization foundation for time series analysis tasks, facilitating the adaptation of sequence modeling skills acquired from vast text data to time series representation learning.

Specifically, we initialized a patchTST and conducted identical two-stage training as \method{} in \shortautoref{sec-method}, encompassing casual next-patch pre-training in the first stage and multi-patch prediction adaption in the second stage. Experimental configurations remained consistent, with the first stage pre-training executed on the Weather, Traffic, Electricity, ILI, and 4 ETT datasets, followed by a representation adaptation and prediction on the 4 ETT datasets. 
Results in \shortautoref{tab:abalation-pre-train} reveal that patchTST without pre-training weights from extensive text data, while showing improvement throughout the two-stage training, still exhibits a discernible disparity compared to \method{} with pre-training weights from extensive text data. This divergence can be attributed to the inadequacy of limited time series data to facilitate learning universal time series representations from the challenging pre-training tasks, whereas the sequence modeling proficiency acquired from text-based LLM pre-training weights offers a solid initialization basis, enabling effective optimization of time series representations despite limited time series data availability. 
This experiment, along with the ablation experiments (B.2 and B.3) in \shortautoref{sec:Ablation_Study} and the interpretability experiment in \shortautoref{sec:Interpretability-Experiment}, collectively emphasize the crucial significance of LLM pre-training weights in learning time series representations, both quantitatively and qualitatively.

\input{tables/abalation-pre-train}

\subsection{Further Exploration Experiment on Patch-wise Decoding}

After PatchTST~\citep{PatchTST} proposed to utilize patches as the foundational unit for time series processing, owing to their capacity to capture local semantic information while minimizing memory usage, the adoption of patches as the core processing unit for time series analysis tasks has become widespread~\citep{onefitsall,SimMTM,independent_patch}. However, previous patch-based models have predominantly focused on devising diverse encoder model architectures, overlooking the design of the final decoding layer. These models typically employ a simple approach: after generating a sequence of patches representation $\{\vp_1, \cdots, \vp_{L_p} \}, \vp_i \in \R^D$ across $L_p$ patches with dimension $D$ from the model backbone, these patches are concatenated and unfolded into a one-dimensional sequence, followed by a large sequence-wise layer $\mathbf{W}_{s} \in \R^{(L_p \cdot D) \times H}$ to decode the patch representation into prediction horizon $H$. Yet, two issues necessitate attention: (1) \textbf{the extensive linear decoding layer might concentrate the primary gradient updates of the model}, leading to insufficient training of the model backbone and overfitting risks to downstream tasks; (2) \textbf{the fully connected linear layer fails to disentangle the patch encoding and decoding processes}, hindering the optimization of patch representations by the model backbone and exacerbating overfitting of this fully connected layer to downstream tasks. 

In our \method{}, we introduce a patch-wise decoder $\mathbf{W}_{p} \in \R^{D \times P}$ to naturally tackle these challenges, reducing the parameter size to only $\frac{P}{L * H}$ (\emph{e.g.}, $0.34\%$, when $P=16, L_p=64, H=720$) of previous sequence-wise decoding layers while disentangling the patch representation encoding and decoding processes. $P$ denotes the patch size. This empowers the LLM backbone and patch-wise decoder to excel in their respective roles: enhancing the representation of each patch and autonomously decoding each patch into the temporal domain. Additionally, we conduct supplementary experiments to elucidate the deficiencies of previous models with sequence-level decoding, confirming that their model performance primarily overfits to the final large fully connected linear layer. Specifically, by initializing the state-of-the-art time series analysis model based on patches, PatchTST, with all other configurations unchanged except for substituting the final sequence-level decoding linear layer with a patch-wise decoding linear layer, the experimental results in \shortautoref{tab:abalation-patch-wise-decoding} reveal a performance deterioration of over $37.6\%$. This underscores the considerable risk of overfitting associated with the sequence-wise linear layer commonly employed in the past, as well as the inadequate optimization of the model backbone.

\input{tables/abalation-patch-wise-decoding}

\subsection{Further Exploration Experiment on Look-back Window Length}

In the first stage of \method{}, referred to as casual next-patch pre-training, we provide a specified number of patch sequences where each patch casually predicts the next patch. The default look-back window length for this patch sequence is set to $720$. However, considering the long-term forecasting tasks downstream, where the default value of the look-back window length is $336$ and the prediction horizon length is $720$, the sum of these two exceeds the maximum length of the continual casual next-patch pre-training. This inconsistency might cause a drop in performance in downstream tasks due to the extra workload of the look-back window length shift, including inadequate adjustment of position embedding, among other factors.
We extended the look-back window length in the first-stage causal next-patch pre-training from $720$ to $1024$, maintaining consistency with \shortautoref{sec:Long-Term_Time_Series_Forecasting} for other settings. Comparative experiments were performed in downstream tasks involving long-term forecasting and imputation, with results presented in \shortautoref{tab:look-back-window-forecasting} and \shortautoref{tab:look-back-window-imputation}. It is evident that augmenting the look-back window length in the first stage can effectively alleviate inconsistencies in the observable horizon between upstream and downstream tasks of \method{}, thereby improving the model's overall performance, particularly for long-term prediction horizons, such as a prediction horizon of $720$.

\input{tables/look-back-window}

\subsection{Detailed Results for Classification}

To assess the performance of the representations learned by \method{} in high-level abstract representation tasks and their generalizability across datasets, we evaluated \method{} at the sequence level after stage 1 casual next-patch pre-training, using pre-training data consistent with long-term forecasting in \shortautoref{sec:Long-Term_Time_Series_Forecasting}. Specifically, we adopted the experiment settings outlined in GPT4TS: For classification, we utilized 10 multivariate UEA classification datasets~\citep{uea} encompassing gesture, action, audio recognition, medical diagnosis, and other practical tasks. Full classification experiment results are in \shortautoref{tab:classification}.

\input{tables/classification}

\subsection{Detailed Results for Imputation}

We perform experiments on four ETT datasets (ETTh1, ETTh2, ETTm1, ETTm2), each exhibiting common data-missing scenarios. Adhering to the GPT4TS configurations, we assess different random mask ratios ($\{12.5\%, 25\%, 37.5\%, 50\%\}$) for evaluating diverse levels of missing data.

The results, presented in \shortautoref{tab:imputation}, affirm that \method{} excels in performance across the majority of datasets. Notably, when contrasted with the prior state-of-the-art models TimesNet and GPT4TS, \method{} demonstrates more than \textbf{9.19\%} decrease in mean squared error (MSE) for ETTh1 and competitive SOTA performance across the four benchmark datasets. This substantiates the efficacy of the proposed approach in discerning temporal patterns within incomplete time series.

\input{tables/imputation}
\input{tables/imputation_full}

\newpage
\subsection{Detailed Results for Anomaly Detection}
\input{tables/anomaly_full}

%% file: tables/long-term-forecasting-full.tex
\begin{table*}[!h]
% \captionsetup{font=small} 
\caption{\textbf{Full Long-term Forecasting Results.} We use forecasting horizons $H \in \{96, 192, 336, 720\}$. \textcolor{blue}{We calculate the MSE for each dataset.} A lower value indicates better performance. {\boldres{Red}}: the best, \secondres{Underlined}: the second best.}
\label{tab:long-term-forecasting-full}
% \vspace{-5mm}
\begin{center}
\begin{small}
\scalebox{0.8}{
\setlength{\tabcolsep}{1.0mm}{
\begin{tabular}{c|c|c|c|c|c|c|c|c|c|c|c|c|c|c}
\toprule
\multicolumn{2}{c|}{Methods} & \method{} & GPT4TS & Time-LLM & DLinear & PatchTST & TimesNet & FEDformer & Autoformer & Stationary & ETSformer & LightTS & Informer & Reformer \\
\midrule

\multirow{4}{*}{\rotatebox{90}{Weather}}
& 96  & \boldres{0.149}   & 0.162 & 0.163 & 0.176 & \secondres{0.152} & 0.172 & 0.217 & 0.266 & 0.173 & 0.197 &  0.182 & 0.300 & 0.689 \\
& 192 &\boldres{0.190}   & 0.204 & 0.206 & 0.220 & \secondres{0.197}  & 0.219 & 0.276 & 0.307 & 0.245 & 0.237 & 0.227 & 0.598 & 0.752 \\
& 336 &\boldres{0.238}   & 0.254 & 0.255 & 0.265 & \secondres{0.249} & 0.280 & 0.339 & 0.359 & 0.321 & 0.298 & 0.282 & 0.578 & 0.639 \\
& 720 & \boldres{0.316}   & 0.326 & 0.325 & 0.333 & \secondres{0.320} & 0.365 & 0.403 & 0.419 & 0.414 & 0.352 & 0.352 & 1.059 & 1.130 \\
\midrule

\multirow{4}{*}{\rotatebox{90}{Traffic}}
& 96  & \secondres{0.372}   & 0.388 & 0.383 & 0.410 & \boldres{0.367} &  0.593 & 0.587 & 0.613 & 0.612 & 0.607 & 0.615 & 0.719 & 0.732 \\
& 192 &\boldres{0.383}   &  0.407 & 0.398 &  0.423 & \secondres{0.385} & 0.617 & 0.604 & 0.616 & 0.613 & 0.621 & 0.601 & 0.696 & 0.733 \\
& 336 &\boldres{0.396}   & 0.412  & 0.407 & 0.436 & \secondres{0.398} & 0.629 & 0.621 & 0.622 & 0.618 & 0.622 & 0.613 & 0.777 & 0.742 \\
& 720 & \boldres{0.434}   & 0.450 & \secondres{0.434} & 0.466 & \secondres{0.434} & 0.640  & 0.626 & 0.660 & 0.653 & 0.632 & 0.658 & 0.864 & 0.755 \\
\midrule

\multirow{4}{*}{\rotatebox{90}{Electricity}}
& 96  & \boldres{0.127}   & 0.139 & 0.140 & 0.140 & \secondres{0.130} & 0.168 & 0.193 & 0.201 & 0.169 & 0.187 & 0.207 & 0.274 & 0.312\\
& 192 &\boldres{0.145}   & 0.153  & 0.151 & 0.153 & \secondres{0.148} & 0.184 & 0.201 & 0.222 & 0.182 & 0.199 & 0.213 & 0.296 & 0.348 \\
& 336 &\boldres{0.163}   & 0.169 & 0.171 & 0.169 & \secondres{0.167} & 0.198 & 0.214 & 0.231 & 0.200 & 0.212 & 0.230 & 0.300 & 0.350\\
& 720 & \secondres{0.206}   & 0.206 & 0.210 & 0.203 & \boldres{0.202} & 0.220 & 0.246 & 0.254 & 0.222 & 0.233 & 0.265 & 0.373 & 0.340\\
\midrule

\multirow{4}{*}{\rotatebox{90}{$ETTh1$}}
& 96  &\secondres{0.380} & 0.376 & 0.399 & \boldres{0.375} &\boldres{0.375} &0.384&0.376&0.449&0.513&0.494&0.424&0.865&0.837\\
& 192 &\boldres{0.396}  & 0.416 & 0.433 & \secondres{0.405}  & 0.414 &0.436&0.420&0.500&0.534&0.538&0.475&1.008&0.923\\
& 336 &\boldres{0.413} & 0.442 & 0.469 &0.439& \secondres{0.431} &0.491&0.459&0.521&0.588&0.574&0.518&1.107&1.097\\
& 720 & \secondres{0.461}  & 0.477 & 0.473 & 0.472  & \boldres{0.449} &0.521&0.506&0.514&0.643&0.562&0.547&1.181&1.257\\
\midrule

\multirow{4}{*}{\rotatebox{90}{$ETTh2$}}
& 96  &\boldres{0.251}  & 0.285 & 0.294 & 0.289  & \secondres{0.274}&0.340&0.358&0.346&0.476&0.340&0.397&3.755&2.626\\
& 192 &\boldres{0.298}  & 0.354 & 0.355 & 0.383 & \secondres{0.339} &0.402&0.429&0.456&0.512&0.430&0.520&5.602&11.12\\
& 336 &\secondres{0.343} & 0.373 & 0.372 & 0.448  & \boldres{0.331} &0.452&0.496&0.482&0.552&0.485&0.626&4.721&9.323 \\
& 720 & 0.417  & \secondres{0.406} & 0.428 & 0.605  & \boldres{0.379} &0.462&0.463&0.515&0.562&0.500&0.863&3.647&3.874\\
% & Avg &\boldres{0.327}&0.381&0.431&\secondres{0.330}&0.414&0.437&0.450&0.526&0.439&0.602&4.431&6.736\\
\midrule

\multirow{4}{*}{\rotatebox{90}{$ETTm1$}}
& 96  &\boldres{0.290}  & 0.292 & 0.293 & 0.299  & \secondres{0.290} &0.338&0.379&0.505&0.386&0.375&0.374&0.672&0.538 \\
& 192 &\boldres{0.316} & 0.332 & 0.333 & 0.335  & \secondres{0.332} &0.374&0.426&0.553&0.459&0.408&0.400&0.795&0.658\\
& 336 &\boldres{0.342} & \secondres{0.366} & 0.367 & 0.369  & 0.366 &0.410&0.445&0.621&0.495&0.435&0.438&1.212&0.898\\
& 720 &\boldres{0.381}  & \secondres{0.417} & 0.435 & 0.425  & 0.420 &0.478&0.543&0.671&0.585&0.499&0.527&1.166&1.102\\
\midrule

\multirow{4}{*}{\rotatebox{90}{$ETTm2$}}
& 96  &0.171  & 0.173 & 0.178 & \secondres{0.167}  & \boldres{0.165} &0.187&0.203&0.255&0.192&0.189&0.209&0.365&0.658\\
& 192 & 0.234  & 0.229 & 0.245 & \secondres{0.224}  & \boldres{0.220} &0.249&0.269&0.281&0.280&0.253&0.311&0.533&1.078\\
& 336 & 0.291  & 0.286 & 0.298 & \secondres{0.281}  & \boldres{0.278}  &0.321&0.325&0.339&0.334&0.314&0.442&1.363&1.549\\
& 720 & 0.393  & \secondres{0.378} & 0.393 & 0.397  & \boldres{0.367} &0.408&0.421&0.433&0.417&0.414&0.675&3.379&2.631 \\
\midrule

\multirow{4}{*}{\rotatebox{90}{$ILI$}}
& 24  & \boldres{1.359}  & 2.063 & 1.617 & 2.215  & \secondres{1.522} &2.317& 3.228 & 
3.483 &2.294 & 2.527 & 8.313 & 5.764 & 4.400 \\
& 36 & \boldres{1.405}  & 1.868 & 1.708 & 1.963  & \secondres{1.430} & 1.972 & 2.679 &3.103 &1.825 &2.615 &6.631 &4.755 & 4.783\\
& 48 & \boldres{1.442}  & 1.790 & 1.633 & 2.130  &  \secondres{1.673} & 2.238 & 2.622 & 2.669 &2.010 &2.359 &7.299 &4.763&4.832\\
& 60 & \secondres{1.603}  & 1.979 & 2.106 & 2.368  & \boldres{1.529} &2.027&2.857&2.770&2.178&2.487&7.283&5.264&4.882 \\

\bottomrule
\end{tabular}
}
}
\end{small}
\end{center}
\end{table*}

%% file: tables/few-shot-forecasting.tex
\begin{table*}[t]
\caption{\textbf{Full Few-shot Learning Results on 5\% Data.} We use prediction length $O \in \{96, 192, 336, 720\}$. A lower MSE indicates better performance, and the best results are highlighted in \boldres{red}. '-' means that 5\% time series is not sufficient to constitute a training set.}
\label{tab:few-shot-forecasting-5per}
% \vskip 0.15in
\begin{center}
\begin{small}
\scalebox{0.8}{
\setlength{\tabcolsep}{0.6mm}{
\begin{tabular}{c|c|cc|cc|cc|cc|cc|cc|cc|cc|cc|cc|cc|cc}
\toprule

\multicolumn{2}{c|}{Methods}&\multicolumn{2}{c|}{\method{}}&\multicolumn{2}{c|}{GPT4TS}&\multicolumn{2}{c|}{DLinear}&\multicolumn{2}{c|}{PatchTST}&\multicolumn{2}{c|}{TimesNet}&\multicolumn{2}{c|}{FEDformer}&\multicolumn{2}{c|}{Autoformer}&\multicolumn{2}{c|}{Stationary}&\multicolumn{2}{c|}{ETSformer}&\multicolumn{2}{c|}{LightTS}&\multicolumn{2}{c|}{Informer}&\multicolumn{2}{c}{Reformer} \\

\midrule

\multicolumn{2}{c|}{Metric} & MSE  & MAE & MSE & MAE& MSE & MAE& MSE  & MAE& MSE  & MAE& MSE  & MAE& MSE  & MAE& MSE  & MAE& MSE  & MAE& MSE  & MAE& MSE  & MAE& MSE  & MAE\\

\midrule

\multirow{5}{*}{\rotatebox{90}{$ETTh1$}}
& 96  & 0.553 & 0.479  & 0.543 & 0.506 & 0.547 & 0.503 & 0.557 & 0.519 & 0.892& 0.625& 0.593 & 0.529 & 0.681 & 0.570 &0.952&0.650& 1.169 & 0.832 & 1.483 & 0.91 & 1.225 & 0.812 &1.198&0.795 \\
& 192 & 0.597 & 0.498   & 0.748 & 0.580 & 0.720 & 0.604 & 0.711 & 0.570 &0.940 & 0.665& 0.652 & 0.563 & 0.725 & 0.602 &0.943&0.645& 1.221 & 0.853 & 1.525 & 0.93 & 1.249 & 0.828 &1.273&0.853\\
& 336 & 0.674 & 0.544   & 0.754 & 0.595 & 0.984 & 0.727 & 0.816 & 0.619 & 0.945&0.653 & 0.731 & 0.594 & 0.761 & 0.624 &0.935&0.644& 1.179 & 0.832 & 1.347 & 0.87 & 1.202 & 0.811 &1.254&0.857 \\
& 720 & - & - & - & - & - & - & - & - & - & - & - & - & - & - & - & - & - & - & - & - & - & - & - & - \\
&Avg.& \boldres{0.608} &\boldres{0.507}&0.681&0.560&0.750&0.611&0.694&0.569&0.925&0.647&0.658&0.562&0.722&0.598&0.943&0.646&1.189&0.839&1.451&0.903&1.225&0.817&1.241&0.835\\
\midrule

\multirow{5}{*}{\rotatebox{90}{$ETTh2$}}
& 96  & 0.331 & 0.392  & 0.376 & 0.421 & 0.442 & 0.456 & 0.401 & 0.421 & 0.409& 0.420& 0.390 & 0.424 & 0.428 & 0.468 &0.408&0.423 & 0.678 & 0.619 & 2.022 & 1.006 & 3.837 & 1.508 &3.753&1.518\\
& 192 & 0.374 & 0.417 & 0.418 & 0.441 & 0.617 & 0.542 & 0.452 & 0.455 & 0.483& 0.464& 0.457 & 0.465 & 0.496 & 0.504 &0.497&0.468 & 0.845 & 0.697 & 3.534 & 1.348 & 3.975 & 1.933 &3.516&1.473\\
& 336 & 0.418 & 0.443 & 0.408 & 0.439 & 1.424 & 0.849 & 0.464 & 0.469 & 0.499& 0.479& 0.477 & 0.483 & 0.486 & 0.496 &0.507&0.481 & 0.905 & 0.727 & 4.063 & 1.451 & 3.956 & 1.520 &3.312&1.427\\
& 720 & - & - & - & - & - & - & - & - & - & - & - & - & - & - & - & - & - & - & - & - & - & - & - & - \\
&Avg.& \boldres{0.374} & \boldres{0.417}&0.400&0.433&0.827&0.615&0.439&0.448&0.463&0.454&0.441&0.457&0.47&0.489&0.470&0.457&0.809&0.681&3.206&1.268&3.922&1.653&3.527&1.472\\
\midrule

\multirow{5}{*}{\rotatebox{90}{$ETTm1$}}
& 96  & 0.372 & 0.387  & 0.386 & 0.405 & 0.332 & 0.374 & 0.399 & 0.414 & 0.606& 0.518& 0.628 & 0.544 & 0.726 & 0.578 &0.823&0.587 & 1.031 & 0.747 & 1.048 & 0.733 & 1.130 & 0.775 &1.234&0.798\\
& 192 & 0.407 & 0.405  & 0.440 & 0.438 & 0.358 & 0.390 & 0.441 & 0.436 & 0.681& 0.539& 0.666 & 0.566 & 0.750 & 0.591 &0.844&0.591 & 1.087 & 0.766 & 1.097 & 0.756 & 1.150 & 0.788 &1.287&0.839\\
& 336 & 0.433 & 0.421  & 0.485 & 0.459 & 0.402 & 0.416 & 0.499 & 0.467 & 0.786& 0.597& 0.807 & 0.628 & 0.851 & 0.659 &0.870&0.603 & 1.138 & 0.787 & 1.147 & 0.775 & 1.198 & 0.809 &1.288&0.842\\
& 720 & 0.464 & 0.442 & 0.577 & 0.499 & 0.511 & 0.489 & 0.767 & 0.587 & 0.796& 0.593& 0.822 & 0.633 & 0.857 & 0.655 &0.893&0.611 & 1.245 & 0.831 & 1.200 & 0.799 & 1.175 & 0.794 &1.247&0.828\\
&Avg.&0.419&\boldres{0.414}&0.472&0.450&\boldres{0.400}&0.417&0.526&0.476&0.717&0.561&0.730&0.592&0.796&0.620&0.857&0.598&1.125&0.782&1.123&0.765&1.163&0.791&1.264&0.826\\
\midrule

\multirow{5}{*}{\rotatebox{90}{$ETTm2$}}
& 96  & 0.214 & 0.293 & 0.199 & 0.280 & 0.236 & 0.326 & 0.206 & 0.288 & 0.220& 0.299& 0.229 & 0.320 & 0.232 & 0.322 &0.238&0.316 & 0.404 & 0.485 & 1.108 & 0.772 & 3.599 & 1.478&3.883&1.545 \\
& 192 & 0.268 & 0.322  & 0.256 & 0.316 & 0.306 & 0.373 & 0.264 & 0.324 & 0.311& 0.361& 0.394 & 0.361 & 0.291 & 0.357 &0.298&0.349 & 0.479 & 0.521 & 1.317 & 0.850 & 3.578 & 1.475 &3.553&1.484\\
& 336 & 0.305 & 0.355 & 0.318 & 0.353 & 0.380 & 0.423 & 0.334 & 0.367 & 0.338& 0.366& 0.378 & 0.427 & 0.478 & 0.517 &0.353&0.380 & 0.552 & 0.555 & 1.415 & 0.879 & 3.561 & 1.473 &3.446&1.460\\
& 720 & 0.401 & 0.412  & 0.460 & 0.436 & 0.674 & 0.583 & 0.454 & 0.432 & 0.509& 0.465& 0.523 & 0.510 & 0.553 & 0.538 &0.475&0.445 & 0.701 & 0.627 & 1.822 & 0.984 & 3.896 & 1.533 &3.445&1.460\\
&Avg.&\boldres{0.297}&\boldres{0.345}&0.308&0.346&0.399&0.426&0.314&0.352&0.344&0.372&0.381&0.404&0.388&0.433&0.341&0.372&0.534&0.547&1.415&0.871&3.658&1.489&3.581&1.487\\

\bottomrule
\end{tabular}
}
}
\end{small}
\end{center}
\end{table*}

%% file: tables/short-term-forecasting-full.tex
\begin{table*}[ht]
\renewcommand\arraystretch{1.5}
\caption{\textbf{Full Results of Short-term Forecasting.} A lower value indicates better performance. {\boldres{Red}}: the best.}
\label{tab:short_term_full}
%\vskip 0.15in
\begin{center}
\begin{small}
\scalebox{0.8}{
\setlength{\tabcolsep}{0.55mm}{
\begin{tabular}{cc|cccccccccccccc}
\toprule

\multicolumn{2}{c|}{Methods}&\method{}&GPT4TS&TimesNet&PatchTST&N-HiTS&N-BEATS& ETSformer& LightTS& DLinear &FEDformer &Stationary &Autoformer  &Informer&Reformer \\

\midrule
\multirow{3}{*}{\rotatebox{90}{$Yearly$}}
&SMAPE& 13.540  &14.804&\boldres{13.387}&13.477&13.418&13.436&18.009&14.247&16.965&13.728&13.717&13.974&14.727&16.169\\
&MASE& 3.108 &3.608&\boldres{2.996}&3.019&3.045&3.043&4.487&3.109&4.283&3.048&3.078&3.134&3.418&3.800\\
&OWA& 0.805  &0.907&\boldres{0.786}&0.792&0.793&0.794&1.115&0.827&1.058&0.803&0.807&0.822&0.881&0.973\\
\bottomrule

\multirow{3}{*}{\rotatebox{90}{$Quarterly$}}
&SMAPE& 10.216 &10.508&\boldres{10.100}&10.38&10.202&10.124&13.376&11.364&12.145&10.792&10.958&11.338&11.360&13.313\\
&MASE& 1.1971 &1.233&\boldres{1.182}&1.233&1.194&1.169&1.906&1.328&1.520&1.283&1.325&1.365&1.401&1.775\\
&OWA& 0.900 &0.927&\boldres{0.890}&0.921&0.899&0.886&1.302&1.000&1.106&0.958&0.981&1.012&1.027&1.252\\
\bottomrule

\multirow{3}{*}{\rotatebox{90}{$Monthly$}}
&SMAPE& 12.775 &12.981&\boldres{12.670}&12.959&12.791&12.677&14.588&14.014&13.514&14.260&13.917&13.958&14.062&20.128\\
&MASE& 0.944 &0.956&\boldres{0.933}&0.970&0.969&0.937&1.368&1.053&1.037&1.102&1.097&1.103&1.141&2.614\\
&OWA& 0.887 &0.899&\boldres{0.878}&0.905&0.899&0.880&1.149&0.981&0.956&1.012&0.998&1.002&1.024&1.927\\
\bottomrule

\multirow{3}{*}{\rotatebox{90}{$Others$}}
&SMAPE& 5.032 &5.282&\boldres{4.891}&4.952&5.061&4.925&7.267&15.880&6.709&4.954&6.302&5.485&24.460&32.491\\
&MASE& 3.481 &3.573&3.302&3.347&\boldres{3.216}&3.391&5.240&11.434&4.953&3.264&4.064&3.865&20.960&33.355\\
&OWA& 1.078 &1.119&\boldres{1.035}&1.049&1.040&1.053&1.591&3.474&1.487&1.036&1.304&1.187&5.879&8.679\\
\bottomrule

\multirow{3}{*}{\rotatebox{90}{$Average$}}
&SMAPE& 11.950
&12.422 &\boldres{11.829}&12.059& 11.927& 11.851& 14.718& 13.525& 13.639 &12.840 &12.780 &12.909 &14.086 &18.200 \\
&MASE& 1.629
& 1.763 & \boldres{1.585}&1.623 & 1.613 & 1.599 &2.408 &2.111 &2.095 &1.701 &1.756 &1.771  &2.718 &4.223\\
&OWA& 0.867
&0.919 & \boldres{0.851}&0.869 &0.861 &0.855 &1.172 &1.051 &1.051 &0.918 &0.930 &0.939 & 1.230 & 1.775\\

\bottomrule

\end{tabular}
}
}
\end{small}
\end{center}
\end{table*}

%% file: tables/long-term-forecasitng-with-std.tex
\begin{table}[ht]
  \caption{\textbf{Standard Deviations of Our \method{} and the Second-Best Method (PatchTST)} on ETTh1, ETTh2, ETTm1, and ETTm2 for Long-term Forecasting.}
  \label{tab:errorbar_long}
  % \vskip 0.05in
  \centering
  \begin{small}
  \renewcommand{\multirowsetup}{\centering}
  \begin{tabular}{l|cc|cc}
    \toprule
    Model & \multicolumn{2}{c|}{\method{}} & \multicolumn{2}{c}{PatchTST \citeyearpar{PatchTST}}  \\
    \cmidrule(lr){0-1}\cmidrule(lr){2-3}\cmidrule(lr){4-5}
    Dataset & MSE & MAE & MSE & MAE\\
    \midrule
    ETTh1 & $0.413 \pm 0.008$ & $0.429 \pm 0.007$ &$0.413 \pm 0.002$ &$0.431 \pm 0.003$  \\
    ETTh2 & $0.327 \pm 0.006$ & $0.385 \pm 0.004$ & $0.330 \pm 0.002$ & $0.379 \pm 0.007$  \\
    ETTm1 & $0.332 \pm 0.006$ & $0.386 \pm 0.009$ &$0.352 \pm 0.006$ & $0.382 \pm 0.004$ \\
    ETTm2 & $0.271 \pm 0.003$ & $0.332 \pm 0.005$ &$0.258 \pm 0.003$ & $0.315 \pm 0.002$ \\
    \bottomrule
  \end{tabular}
    \end{small}
\end{table}

%% file: tables/zero-classical.tex
\begin{table}[ht]
\caption{\textbf{Comparison with Traditional Methods.} A lower value indicates better performance. {\boldres{Red}}: the best, \secondres{Underlined}: the second best.}
\label{tab:classical_methods}
\begin{center}
\begin{small}
\scalebox{1.0}{
\begin{tabular}{c|c|cc|cc|cc|cc|cc}
\toprule

\multicolumn{2}{c|}{Methods}&\multicolumn{2}{c|}{\method{} 5\%}&\multicolumn{2}{c|}{ETS}&\multicolumn{2}{c|}{ARIMA}&\multicolumn{2}{c}{NaiveDrift}\\

\multicolumn{2}{c|}{Metric} & MSE  & MAE & MSE  & MAE& MSE  & MAE& MSE  & MAE\\

\midrule

\multirow{2}{*}{ETTh2}
& 96  & \boldres{0.331} & \boldres{0.392}  & 2.954  & 0.742 & \secondres{0.481} & \secondres{0.443} & 0.764 & 0.561 \\
& 192 & \boldres{0.374} & \boldres{0.417}  & 10.226 & 1.212 & 0.585 & 0.495 & 1.560 & 0.785 \\
\midrule

\multirow{2}{*}{ETTm1}
& 96  & \boldres{0.413} & \boldres{0.397} & 52.237 & 2.689 & \secondres{0.693} & \secondres{0.547} & 1.539  & 0.913 \\
& 192 & \boldres{0.416} & \boldres{0.399} & 186.445 & \secondres{4.654} & \secondres{0.710} & 0.557 & 2.869 & 1.215 \\
\bottomrule
\end{tabular}
}
\end{small}
\end{center}
\end{table}

%% file: tables/abalation-full.tex
\begin{table*}[h!]
\caption{\textbf{Full Ablation Results on 4 ETT Datasets in Long-term Forecasting.} {\boldres{Red}}: the best, \secondres{Underlined}: the second best.}
\label{tab:abalation-full}
% \vskip 0.15in
\begin{center}
\begin{small}
\scalebox{0.56}{
\setlength{\tabcolsep}{0.5mm}{
\begin{tabular}{c|c|cc|cc|cc|cc|cc|cc|cc|cc|cc|cc|cc|cc|cc}
\toprule

\multicolumn{2}{p{2.0cm}|}{Methods}&\multicolumn{2}{p{2.0cm}|}{\makecell{\textbf{A.1} \\ \method{}}}&\multicolumn{2}{p{2.0cm}|}{\makecell{\textbf{A.2} \\ \method{}~(3)}}&\multicolumn{2}{p{2.0cm}|}{\makecell{\textbf{A.3} \\ \method{}~(9)}}&\multicolumn{2}{p{2.0cm}|}{\makecell{\textbf{A.4} \\ \method{}~(12)}}&\multicolumn{2}{p{2.0cm}|}{\makecell{\textbf{B.1} \\ w/o Casual \\ Continual \\ Pretraining}}&\multicolumn{2}{p{2.0cm}|}{\makecell{\textbf{B.2} \\ w/o LLM \\ Pretrained \\ Weights~(6)}}&\multicolumn{2}{p{2.0cm}|}{\makecell{\textbf{B.3} \\ w/o LLM \\ Pretrained \\ Weights~(3)}}&\multicolumn{2}{p{2.0cm}|}{\makecell{\textbf{C.1} \\ w/o Patch-level \\ Decoder}}&\multicolumn{2}{p{2.0cm}|}{\makecell{\textbf{C.2} \\ w/o Position \\-aware \\ Attention \\ Mask}}&\multicolumn{2}{p{2.0cm}|}{\makecell{\textbf{D.1}  \\ Init with FFT}}&\multicolumn{2}{p{2.0cm}|}{\makecell{\textbf{D.2} \\ Init with \\ Random}}&\multicolumn{2}{p{2.0cm}|}{\makecell{\textbf{E.1} \\ LN+PE+Attn}}&\multicolumn{2}{p{2.0cm}}{\makecell{\textbf{E.2} \\ LN+PE+ \\ Attn+FFN}} \\

\midrule

\multicolumn{2}{c|}{Metric} & MSE  & MAE & MSE  & MAE & MSE & MAE& MSE & MAE& MSE  & MAE& MSE  & MAE& MSE  & MAE& MSE  & MAE& MSE  & MAE& MSE  & MAE& MSE  & MAE& MSE  & MAE& MSE  & MAE\\
\midrule

\multirow{5}{*}{\rotatebox{90}{$ETTh1$}}
& 96  & 0.380 & 0.407 & 0.410 & 0.437 & 0.412 & 0.436 & 0.575 & 0.527 & 0.444 & 0.459 & 0.410 & 0.435 & 0.432 & 0.455 & 0.412 & 0.425 & 0.419 & 0.435 & 0.380 & 0.406 & 0.378 & 0.406 & 0.408 & 0.442 & 0.424 & 0.440\\
& 192 & 0.396 & 0.417 & 0.419 & 0.444 & 0.434 & 0.448 & 0.573 & 0.530 & 0.450 & 0.466 & 0.419 & 0.442 & 0.444 & 0.464 & 0.447 & 0.445 & 0.435 & 0.437 & 0.407 & 0.422 & 0.415 & 0.425 & 0.424 & 0.451 & 0.444 & 0.451\\
& 336 & 0.413 & 0.428 & 0.429 & 0.453 & 0.452 & 0.461 & 0.570 & 0.535 & 0.450 & 0.473 & 0.431 & 0.453 & 0.455 & 0.475 & 0.457 & 0.463 & 0.438 & 0.455 & 0.421 & 0.434 & 0.458 & 0.448  & 0.439 & 0.461 & 0.463 & 0.463\\
& 720 & 0.461 & 0.462 & 0.488 & 0.490 & 0.508 & 0.495 & 0.602 & 0.558 & 0.496 & 0.506 & 0.487 & 0.489 & 0.515 & 0.514 & 0.502 & 0.496 & 0.481 & 0.487 & 0.454 & 0.461 & 0.537 & 0.504 & 0.496 & 0.495 & 0.527 & 0.500\\
&Avg.& \boldres{0.413} & \boldres{0.429} & 0.437 & 0.456 & 0.452 & 0.460 & 0.580 & 0.533 & 0.460 & 0.476 & 0.437 & 0.455 & 0.462 & 0.477 & 0.455 & 0.457 & 0.443 & 0.454 & \secondres{0.416} & \secondres{0.431} & 0.447 & 0.446 & 0.442 & 0.462 &0.465 & 0.464\\
\midrule

\multirow{5}{*}{\rotatebox{90}{$ETTh2$}}
& 96  & 0.251 & 0.332  & 0.259 & 0.335 & 0.274 & 0.344 & 0.273 & 0.344 & 0.292 & 0.360 & 0.263 & 0.342 & 0.266 & 0.345 & 0.330 & 0.386 &0.297 & 0.365& 0.274 & 0.347 & 0.281 & 0.353 & 0.275 & 0.345 & 0.272 & 0.341 \\
& 192 & 0.298 & 0.363  & 0.305 & 0.367 & 0.319 & 0.374 & 0.313 & 0.371 & 0.329 & 0.386 & 0.303 & 0.363 & 0.310 & 0.376 & 0.400 & 0.426 & 0.330 & 0.388 & 0.323 & 0.379 & 0.325 & 0.380 & 0.320 & 0.374 & 0.319 & 0.371\\
& 336 & 0.343 & 0.396 & 0.344 & 0.398 & 0.348 & 0.399 & 0.339 & 0.393 & 0.352 & 0.407 & 0.352 & 0.402 & 0.355 & 0.413 & 0.380 & 0.425 & 0.351 & 0.407& 0.369 & 0.413 & 0.362 & 0.408 & 0.358 & 0.404 & 0.363 & 0.404\\
& 720 & 0.417 & 0.450 & 0.411 & 0.444 & 0.414 & 0.446 & 0.411 & 0.442 & 0.427 & 0.456 & 0.427 & 0.458 & 0.423 & 0.459 & 0.438 & 0.461 & 0.452 & 0.466 & 0.454 & 0.470 & 0.437 & 0.459 & 0.429 & 0.453 & 0.438 & 0.458\\
&Avg.& \boldres{0.327} & \boldres{0.385} & \secondres{0.330} & \secondres{0.386} & 0.339 & 0.391 & 0.334 & 0.388 & 0.350 & 0.402 & 0.336 & 0.391 & 0.339 & 0.398 & 0.387 & 0.425 & 0.358 & 0.407 & 0.355 & 0.402 & 0.351 & 0.400 & 0.346 & 0.394 & 0.348 & 0.394\\
\midrule

\multirow{5}{*}{\rotatebox{90}{$ETTm1$}}
& 96  & 0.290 & 0.361 & 0.287 & 0.361 & 0.341 & 0.391 & 0.341 & 0.393 & 0.314 & 0.367 & 0.307 & 0.365 & 0.311 & 0.370 & 0.301 & 0.357  & 0.342 & 0.393 & 0.321 & 0.373 & 0.308 & 0.364 & 0.300 & 0.360 & 0.301 & 0.358\\
& 192 & 0.316 & 0.376 & 0.327 & 0.375 & 0.381 & 0.412 & 0.379 & 0.413 & 0.350 & 0.389 & 0.342 & 0.385 & 0.348 & 0.391 & 0.344 & 0.384 & 0.381 & 0.416 & 0.352 & 0.392 & 0.341 & 0.384 & 0.337 & 0.383 & 0.334 & 0.379 \\
& 336 & 0.342 & 0.391 & 0.366 & 0.397 & 0.420 & 0.432 & 0.417 & 0.433 & 0.387 & 0.409 & 0.377 & 0.404 & 0.387 & 0.412 & 0.375 & 0.403 & 0.413 & 0.433 & 0.387 & 0.411 & 0.375 & 0.403 & 0.376 & 0.403 & 0.370 & 0.399\\
& 720 & 0.381 & 0.414 & 0.421 & 0.427 & 0.475 & 0.459 & 0.474 & 0.460 & 0.441 & 0.438 & 0.431 & 0.433 & 0.444 & 0.441 & 0.427 & 0.431 & 0.459 &  0.456 & 0.441 & 0.440 & 0.434 & 0.433 & 0.438 & 0.434 & 0.427 & 0.428\\
&Avg.& \boldres{0.332} & \boldres{0.386} & \secondres{0.350} & \secondres{0.390} & 0.404 & 0.424 & 0.403 & 0.425 & 0.373 & 0.401 & 0.364 & 0.397 & 0.373 & 0.404 & 0.362 & 0.394 & 0.399 & 0.425 & 0.375 & 0.404 & 0.365 & 0.396 & 0.363 & 0.395 & 0.358 & 0.391 \\
\midrule

\multirow{5}{*}{\rotatebox{90}{$ETTm2$}}
& 96  & 0.212 & 0.307 & 0.205 & 0.300 & 0.219 & 0.308 & 0.218 & 0.309 & 0.230 & 0.323 & 0.214 & 0.308 & 0.219 & 0.306 & 0.185 & 0.271 & 0.223 & 0.311 & 0.212  & 0.307 & 0.233 & 0.323 & 0.232 & 0.316 &0.208 & 0.302\\
& 192 & 0.263 & 0.337 & 0.255 & 0.330 & 0.274 & 0.342 & 0.271 & 0.339 & 0.276 & 0.350 & 0.267 & 0.336 & 0.278 & 0.347 & 0.257 & 0.316 & 0.273 & 0.341 & 0.269 & 0.338 & 0.279 & 0.349 & 0.288 & 0.347 & 0.268 & 0.342\\
& 336 & 0.310 & 0.365 & 0.316 & 0.369 & 0.327 & 0.375 & 0.322 & 0.370 & 0.321 & 0.376 & 0.319 & 0.366 & 0.326 & 0.376 & 0.311 & 0.359 & 0.320 & 0.370 & 0.319 & 0.369 & 0.322 & 0.374 & 0.338 & 0.377 & 0.324 & 0.377\\
& 720 & 0.393 & 0.414 & 0.391 & 0.410 & 0.409 & 0.426 & 0.402 & 0.418 & 0.400 & 0.421 & 0.403 & 0.415 & 0.397 & 0.414 & 0.383 & 0.403 & 0.398 & 0.416 & 0.403 & 0.416 & 0.401 & 0.419 & 0.417 & 0.423 & 0.406 & 0.426\\
&Avg.& 0.294 & 0.356 & \secondres{0.292} & \secondres{0.352} & 0.307 & 0.363 & 0.303 & 0.359 & 0.307 & 0.368 & 0.301 & 0.356 & 0.305 & 0.361 & \boldres{0.284} & \boldres{0.337} & 0.303 & 0.335 & 0.301 & 0.358 & 0.309 & 0.366 & 0.319 & 0.366 & 0.302& 0.362\\
% \midrule

\bottomrule
\end{tabular}
}
}
\end{small}
\end{center}
\end{table*}

%% file: tables/abalation-pre-train.tex
\begin{table}[ht]
\caption{
Experiment results for special ablation on the role of pre-trained LLM parameters. We use the MSE and MAE as the default metric. A lower value indicates better performance. 
\method{}-GPT2 denotes the our default \method{} with the pre-trained GPT-2 as model backbone.
\method{}-PatchTST denotes the ablated \method{} with the PatchTST as model backbone.
}
\label{tab:abalation-pre-train}
\begin{center}
\begin{small}
\scalebox{0.90}{
\setlength\tabcolsep{3pt}
\begin{tabular}{cc|cc|cc}
\toprule

\multicolumn{2}{c|}{Methods} 
&\multicolumn{2}{c|}{\method{}-GPT2}  
&\multicolumn{2}{c}{\method{}-PatchTST}\\
Datasets&Horizon& MSE & MAE & MSE & MAE\\

\midrule
\multirow{5}{*}{$ETTh1$}
& 96 
& 0.380 & 0.376 & 0.408 & 0.435 \\
& 192
& 0.396  & 0.416 & 0.427 & 0.455 \\
& 336 
& 0.413 & 0.442 & 0.440 & 0.476 \\
& 720 
& 0.461 & 0.477 & 0.500 & 0.517 \\
\midrule

\multirow{5}{*}{$ETTh2$}
& 96 
& 0.251 & 0.285 & 0.317 & 0.380\\
& 192
& 0.298 & 0.354  & 0.339 & 0.396\\
& 336 
& 0.343 & 0.373 & 0.350 & 0.408\\
& 720 
& 0.417 & 0.406 & 0.440 & 0.463\\

\midrule
\multirow{5}{*}{$ETTm1$}
& 96 
& 0.290 & 0.292 & 0.347 & 0.378 \\
& 192
& 0.316 & 0.332 & 0.382 & 0.396 \\
& 336 
& 0.342 & 0.366 & 0.425 & 0.419 \\
& 720 
& 0.381 & 0.417 & 0.490 & 0.455 \\
\midrule

\multirow{5}{*}{$ETTm2$}
& 96 
& 0.171 & 0.173 & 0.244 & 0.327 \\
& 192
& 0.234 & 0.229 & 0.285 & 0.350 \\
& 336 
& 0.291 & 0.286 & 0.327 & 0.374\\
& 720 
& 0.393 & 0.378 & 0.406 & 0.417\\

\bottomrule

\end{tabular}
}
\end{small}
\end{center}
\end{table}

%% file: tables/abalation-patch-wise-decoding.tex
\begin{table}[ht]
\caption{
Experiment results for ablation of patch-wise decoding in PatchTST. We use the MSE and MAE as the default metric. A lower value indicates better performance. 
PatchTST-sd denotes the origin PatchTST with the sequence-wise decoding layer.
PatchTST-pd denotes the PatchTST with our patch-wise decoding layer.
}
\label{tab:abalation-patch-wise-decoding}
\begin{center}
\begin{small}
\scalebox{0.90}{
\setlength\tabcolsep{3pt}
\begin{tabular}{cc|cc|cc}
\toprule

\multicolumn{2}{c|}{Methods} 
&\multicolumn{2}{c|}{PatchTST-sd}  
&\multicolumn{2}{c}{PatchTST-pd}\\
Datasets&Horizon& MSE & MAE & MSE & MAE\\

\midrule
\multirow{5}{*}{$ETTh1$}
& 96 
& 0.375 & 0.399 & 0.498 & 0.489 \\
& 192
& 0.414  & 0.421 & 0.497 & 0.492 \\
& 336 
& 0.431 & 0.436 & 0.496 & 0.498 \\
& 720 
& 0.449 & 0.466 & 0.533 & 0.527 \\
\midrule

\multirow{5}{*}{$ETTh2$}
& 96 
& 0.274 & 0.336 & 0.287 & 0.362\\
& 192
& 0.339 & 0.379  & 0.305 & 0.381\\
& 336 
& 0.331 & 0.380 & 0.335 & 0.396\\
& 720 
& 0.379 & 0.422 & 0.414 & 0.446\\

\midrule
\multirow{5}{*}{$ETTm1$}
& 96 
& 0.290 & 0.342 & 0.586 & 0.517 \\
& 192
& 0.332 & 0.369 & 0.595 & 0.522 \\
& 336 
& 0.255 & 0.392 & 0.606 & 0.528 \\
& 720 
& 0.420 & 0.424 & 0.626 & 0.539 \\
\midrule

\multirow{5}{*}{$ETTm2$}
& 96 
& 0.165 & 0.255 & 0.275 & 0.344 \\
& 192
& 0.220 & 0.292 & 0.305 & 0.361 \\
& 336 
& 0.278 & 0.329 & 0.341 & 0.380\\
& 720 
& 0.367 & 0.385 & 0.420 & 0.423\\

\bottomrule

\end{tabular}
}
\end{small}
\end{center}
\end{table}

%% file: tables/look-back-window.tex
\begin{table}[!ht]
    \centering
    \caption{Experiment results for the influence of longer look-back window in long-term forecasting. We use the MSE as the default metric. A lower value indicates better performance. 
    We use forecasting horizons $H \in \{96, 192, 336, 720\}$.
    \method{}-origin(longer) denotes the \method{} with look-back window length as $720$($1024$) in casual next-patch pre-training.
    }
    \vspace{5pt}
    \scalebox{0.65}{
    \setlength{\tabcolsep}{1.0mm}{
    \begin{tabular}{c|cccccccccccc}
    \toprule 
    Datasets & Weather-96 & Weather-192 & Weather-336 & Weather-720 & ETTh2-96 & ETTh2-192 & ETTh2-336 & ETTh2-720 & ETTm2-96 & ETTm2-192 & ETTm2-336 & ETTm2-720 \\
    \midrule  
    \method{}-origin& 0.149 & 0.190 & 0.238 & 0.316  & 0.251 & 0.298 & 0.343 & 0.417 & 0.171 & 0.234 & 0.291 & 0.403 \\
    \method{}-longer& 0.148 & 0.187 & 0.236 & 0.317  & 0.252 & 0.301 & 0.345 & 0.408 & 0.175 & 0.240 & 0.293 & 0.394 \\
    
    \bottomrule 
    \end{tabular}
    }
    }
    \label{tab:look-back-window-forecasting}
\end{table}

\begin{table}[!ht]
\caption{
Experiment results for the influence of longer look-back window in imputation. We use the MSE and MAE as the default metric. A lower value indicates better performance. 
\method{}-origin(longer) denotes the \method{} with look-back window length as $720$($1024$) in casual next-patch pre-training.
}
\label{tab:look-back-window-imputation}
\begin{center}
\begin{small}
\scalebox{0.90}{
\setlength\tabcolsep{3pt}
\begin{tabular}{cc|cc|cc}
\toprule

\multicolumn{2}{c|}{Methods} 
&\multicolumn{2}{c|}{\method{}-origin}  
&\multicolumn{2}{c}{\method{}-longer}\\
Mask&Ratio& MSE & MAE & MSE & MAE\\

\midrule
\multirow{5}{*}{\rotatebox{90}{$ETTh1$}}
& 12.5\% 
& 0.041 & 0.139 & 0.040 & 0.137 \\
& 25\% 
& 0.056  & 0.161 & 0.055 & 0.157 \\
& 37.5\% 
& 0.073 & 0.182 & 0.072 & 0.181 \\
& 50\% 
& 0.098 & 0.210 & 0.097 & 0.209 \\
% & Avg 
% & \boldres{0.067} & \boldres{0.172} &
% \secondres{0.070}&\secondres{0.175}&{0.078}&{0.187}&0.115&0.224&0.202&0.329&0.284&0.373&0.201&0.306&0.117&0.246&0.094&0.201&0.103&0.214&0.161&0.279&0.122&0.245 \\
\midrule

\multirow{5}{*}{\rotatebox{90}{$ETTh2$}}
& 12.5\% 
& 0.040 & 0.126 & 0.040 & 0.125\\
& 25\% 
& 0.046 & 0.138  & 0.046 & 0.138\\
& 37.5\% 
& 0.051 & 0.147 & 0.052 & 0.147\\
& 50\% 
& 0.060 & 0.159 & 0.060 & 0.158\\
% & Avg 
% & \boldres{0.049} & \boldres{0.142} &
% {0.050}&\secondres{0.145}&\secondres{0.049}&{0.146}&0.065&0.163&0.367&0.436&0.119&0.250&0.142&0.259&0.163&0.279&0.053&0.152&0.055&0.156&0.337&0.452&0.234&0.352 \\

\midrule
\multirow{5}{*}{\rotatebox{90}{$ETTm1$}}
& 12.5\% 
& 0.019 & 0.093 & 0.019 & 0.091 \\
& 25\% 
& 0.025 & 0.104 & 0.024 & 0.102 \\
& 37.5\% 
& 0.032 & 0.118 & 0.031 & 0.115 \\
& 50\% 
& 0.045 & 0.139 & 0.044 & 0.137 \\
% & Avg 
% & {0.031} & {0.113} &
% \secondres{0.029}&\secondres{0.107}&\boldres{0.027}&\boldres{0.107}&0.047&0.140&0.120&0.253&0.104&0.218&0.093&0.206&0.062&0.177&0.036&0.126&0.051&0.150&0.071&0.188&0.055&0.166 \\
\midrule

\multirow{5}{*}{\rotatebox{90}{$ETTm2$}}
& 12.5\% 
& 0.019 & 0.079 & 0.019 & 0.078 \\
& 25\% 
& 0.022 & 0.086 & 0.021 & 0.085 \\
& 37.5\% 
& 0.024 & 0.092 & 0.024 & 0.092\\
& 50\% 
& 0.028 & 0.100 & 0.027 & 0.100\\
% & Avg 
% & \secondres{0.023} & {0.089} &
% \secondres{0.023}&\secondres{0.087}&\boldres{0.022}&\boldres{0.088}&0.029&0.102&0.208&0.327&0.046&0.151&0.096&0.208&0.101&0.215&0.026&0.099&0.029&0.105&0.156&0.292&0.157&0.280 \\

\bottomrule

\end{tabular}
}
\end{small}
\end{center}
\end{table}

%% file: tables/classification.tex
\begin{table}[!h]
    \centering
    \caption{Full results for the classification task. We use the accuracy as default metric. A higher value indicates better performance. 
    \boldres{Red}: the best, \secondres{Underlined}: the second best.
    We compare our \method{} with SOTA LLM-based GPT4TS, CNN-based TimesNet, MLP-based DLinear and Machine-Learning-based XGBoost.}
    \vspace{5pt}
    \scalebox{0.9}{
    \setlength{\tabcolsep}{1.0mm}{
    \begin{tabular}{c|ccccc}
    \toprule 
    Methods & \method{} & GPT4TS & TimesNet & DLinear & XGBoost\\
    \midrule  
    EthanolConcentration & \secondres{33.5} & 31.2 & 32.7 & 32.6 & \boldres{43.7} \\
    FaceDetection & \secondres{68.3} & 68.2 & \boldres{68.6} & 68.0 & 63.3 \\
    Handwriting & \secondres{27.5} & 20.6 & \boldres{32.8} & 27.0  & 15.8 \\
    Heartbeat & \boldres{78.0} & \secondres{76.6} & 75.6 & 75.1 & 73.2 \\
    JapaneseVowels & 96.8 & \boldres{97.6} & \secondres{97.3} & 96.2 & 86.5 \\
    PEMS-SF & 64.7 & 60.7 & \secondres{89.6} & 75.1 & \boldres{98.3} \\
    SelfRegulationSCP1 & \boldres{92.5} & \secondres{90.8} & 87.7 & 87.3 & 84.6 \\
    SelfRegulationSCP2 & \boldres{57.2} & 54.3 & \secondres{56.1} & 50.5 & 48.9 \\
    SpokenArabicDigits & \boldres{99.0} & \boldres{99.0} & \secondres{98.8} & 81.4 & 69.6 \\
    UWaveGestureLibrary & 85.3 & \secondres{85.9} & \boldres{86.9}  & 82.1 & 75.9 \\
    
    \bottomrule 
    \end{tabular}
    }
    }
    \label{tab:classification}
\end{table}

% \begin{table}[!h]
%     \centering
%     \caption{Full results for the classification task. We use the accuracy as default metric. A higher value indicates better performance. 
%     \boldres{Red}: the best, \secondres{Underlined}: the second best.
%     We compare our \method{} with SOTA LLM-based GPT4TS, CNN-based TimesNet, MLP-based DLinear and Machine-Learning-based XGBoost.}
%     \scalebox{0.6}{
%     \setlength{\tabcolsep}{1.0mm}{
%     \begin{tabular}{c|cccccccccc}
%     \toprule  
%     Datasets & EthanolConcentration & FaceDetection & Handwriting & Heartbeat & JapaneseVowels & PEMS-SF & SelfRegulationSCP1 & SelfRegulationSCP2 & SpokenArabicDigits & UWaveGestureLibrary\\
%     \midrule  
%     XGBoost & 43.7 & 63.3 &  &  &   &  &  &  &  & \\
%     DLinear &  &  &  &  &   &  &  &  &  & \\
%     TimesNet &  &  &  &  &   &  &  &  &  & \\
%     GPT4TS &  &  &  &  &   &  &  &  &  & \\
%     \method{} &  &  &  &  &   &  &  &  &  & \\
%     \bottomrule 
%     \end{tabular}
%     }
%     }
%     \label{tab:classification}
% \end{table}

%% file: tables/imputation.tex
\begin{table}[ht]
% \captionsetup{font=small} 
\caption{\textbf{Brief Results for Imputation Task.} We randomly mask \{12.5\%, 25\%, 37.5\%, 50\%\} time points of 96-length time series. The results are averaged from 4 different mask ratios. A lower value indicates better performance. 
{\boldres{Red}}: the best, \secondres{Underlined}: the second best.}
\label{tab:imputation}
%\vskip 0.15in
\begin{center}
\begin{small}
\scalebox{0.7}{
\setlength\tabcolsep{3pt}
\begin{tabular}{c|cc|cc|cc|cc|cc|cc|cc|cc|cc|cc|cc|cc}
\toprule

\multirow{2}{*}{Methods} 
&\multicolumn{2}{c|}{\method{}}
&\multicolumn{2}{c|}{GPT4TS} & \multicolumn{2}{c|}{TimesNet}&\multicolumn{2}{c|}
{PatchTST}&\multicolumn{2}{c|}
{ETSformer}&\multicolumn{2}{c|}{LightTS}&\multicolumn{2}{c|}{DLinear}&\multicolumn{2}{c|}{FEDformer}&\multicolumn{2}{c|}{Stationary}&\multicolumn{2}{c|}{Autoformer}&\multicolumn{2}{c|}{Informer}&\multicolumn{2}{c}{Reformer} \\
&MSE&MAE&MSE&MAE&MSE&MAE&MSE&MAE&MSE&MAE&MSE&MAE&MSE&MAE&MSE&MAE&MSE&MAE&MSE&MAE&MSE&MAE&MSE&MAE \\

\midrule

ETTh1
& \boldres{0.067}  &  \boldres{0.172} &\secondres{0.070}&\secondres{0.175}&{0.078}& {0.187}&0.115 &0.224 & 0.202& 0.329& 0.284& 0.373& 0.201& 0.306& 0.117& 0.246& 0.094& 0.201 &0.103& 0.214& 0.161& 0.279&0.122& 0.245\\
ETTh2
& \boldres{0.049}  & \boldres{0.142} &{0.050}&\secondres{0.145}&\secondres{0.049}& {0.146}&0.065 &0.163 & 0.367& 0.436& 0.119& 0.250& 0.142& 0.259& 0.163& 0.279& 0.053& 0.152& 0.055& 0.156& 0.337& 0.452&0.234& 0.352\\
ETTm1
& {0.031}  & {0.113} 
&\secondres{0.029}&\secondres{0.107}&\boldres{0.027} &\boldres{0.107}&0.047 &0.140 & 0.120& 0.253& 0.104& 0.218& 0.093& 0.206& 0.062& 0.177& 0.036& 0.126&0.051& 0.150&  0.071& 0.188 & 0.055 & 0.166 \\
ETTm2
& \secondres{0.023} & {0.089} &\secondres{0.023}&\secondres{0.087}&\boldres{0.022}& \boldres{0.088}&0.029 &0.102 & 0.208& 0.327& 0.046& 0.151& 0.096& 0.208& 0.101& 0.215& 0.026& 0.099 &0.029& 0.105& 0.156& 0.292 & 0.157 & 0.280\\

\bottomrule

\end{tabular}
}
\end{small}
\end{center}
\end{table}

%% file: tables/imputation_full.tex
\begin{table}[ht]
\caption{
\textbf{Full Results for the Imputation Task.} A lower value indicates better performance. 
{\boldres{Red}}: the best, \secondres{Underlined}: the second best.
}
\label{tab:imputation_full}
\begin{center}
\begin{small}
\scalebox{0.70}{
\setlength\tabcolsep{3pt}
\begin{tabular}{cc|cc|cc|cc|cc|cc|cc|cc|cc|cc|cc|cc|cc}
\toprule

\multicolumn{2}{c|}{Methods} 
&\multicolumn{2}{c|}{\method{}} 
&\multicolumn{2}{c|}{GPT4TS} & \multicolumn{2}{c|}{TimesNet}& \multicolumn{2}{c|}{PatchTST}&\multicolumn{2}{c|}{ETSformer}&\multicolumn{2}{c|}{LightTS}&\multicolumn{2}{c|}{DLinear}&\multicolumn{2}{c|}{FEDformer}&\multicolumn{2}{c|}{Stationary}&\multicolumn{2}{c|}{Autoformer}&\multicolumn{2}{c}{Informer}&\multicolumn{2}{c}{Reformer} \\
Mask&Ratio&MSE&MAE&MSE&MAE&MSE&MAE&MSE&MAE&MSE&MAE&MSE&MAE&MSE&MAE&MSE&MAE&MSE&MAE&MSE&MAE&MSE&MAE&MSE&MAE \\

\midrule
\multirow{5}{*}{\rotatebox{90}{$ETTh1$}}
& 12.5\% 
& 0.041 & 0.139 &
0.043&0.142&0.057&0.159&0.093&0.201&0.126&0.263&0.240&0.345&0.151&0.267&0.070&0.190&0.060&0.165&0.074&0.182&0.114&0.234&0.074&0.194 \\
& 25\% 
& 0.056 & 0.161 &
0.057&0.160&0.069&0.178&0.107&0.217&0.169&0.304&0.265&0.364&0.180&0.292&0.106&0.236&0.080&0.189&0.090&0.203&0.140&0.262&0.102&0.227 \\
& 37.5\% 
& 0.073 & 0.182 &
0.074&0.183&0.084&0.196&0.120&0.230&0.220&0.347&0.296&0.382&0.215&0.318&0.124&0.258&0.102&0.212&0.109&0.222&0.174&0.293&0.135&0.261 \\
& 50\% 
& 0.098 & 0.210 &
0.107&0.216&0.102&0.215&0.141&0.248&0.293&0.402&0.334&0.404&0.257&0.347&0.165&0.299&0.133&0.240&0.137&0.248&0.215&0.325&0.179&0.298 \\
& Avg 
& \boldres{0.067} & \boldres{0.172} &
\secondres{0.070}&\secondres{0.175}&{0.078}&{0.187}&0.115&0.224&0.202&0.329&0.284&0.373&0.201&0.306&0.117&0.246&0.094&0.201&0.103&0.214&0.161&0.279&0.122&0.245 \\
\midrule

\multirow{5}{*}{\rotatebox{90}{$ETTh2$}}
& 12.5\% 
& 0.040 & 0.126 &
0.041&0.129&0.040&0.130&0.057&0.152&0.187&0.319&0.101&0.231&0.100&0.216&0.095&0.212&0.042&0.133&0.044&0.138&0.305&0.431&0.163&0.289 \\
& 25\% 
& 0.046 & 0.138 &
0.046&0.139&0.046&0.141&0.061&0.158&0.279&0.390&0.115&0.246&0.127&0.247&0.137&0.258&0.049&0.147&0.050&0.149&0.322&0.444&0.206&0.331 \\
& 37.5\% 
& 0.051 & 0.147 &
0.053& 0.150&0.052&0.151&0.067&0.166&0.400&0.465&0.126&0.257&0.158&0.276&0.187&0.304&0.056&0.158&0.060&0.163&0.353&0.462&0.252&0.370 \\
& 50\% 
& 0.060 & 0.159 &
0.060&0.160&0.060&0.162&0.073&0.174&0.602&0.572&0.136&0.268&0.183&0.299&0.232&0.341&0.065&0.170&0.068&0.173&0.369&0.472&0.316&0.419 \\
& Avg 
& \boldres{0.049} & \boldres{0.142} &
{0.050}&\secondres{0.145}&\secondres{0.049}&{0.146}&0.065&0.163&0.367&0.436&0.119&0.250&0.142&0.259&0.163&0.279&0.053&0.152&0.055&0.156&0.337&0.452&0.234&0.352 \\

\midrule
\multirow{5}{*}{\rotatebox{90}{$ETTm1$}}
& 12.5\% 
& 0.019 & 0.093 &
0.018&0.089&0.023&0.101&0.041&0.130&0.096&0.229&0.093&0.206&0.080&0.193&0.052&0.166&0.032&0.119&0.046&0.144&0.063&0.180&0.042&0.146 \\
& 25\% 
& 0.025 & 0.104 &
0.023&0.099&0.023&0.101&0.044&0.135&0.096&0.229&0.093&0.206&0.080&0.193&0.052&0.166&0.032&0.119&0.046&0.144&0.063&0.180&0.042&0.146 \\
& 37.5\% 
& 0.032 & 0.118 &
0.030&0.111&0.029&0.111&0.049&0.143&0.133&0.271&0.113&0.231&0.103&0.219&0.069&0.191&0.039&0.131&0.057&0.161&0.079&0.200&0.063&0.182 \\
& 50\% 
& 0.045 & 0.139 &
0.041&0.130&0.036&0.124&0.055&0.151&0.186&0.323&0.134&0.255&0.132&0.248&0.089&0.218&0.047&0.145&0.067&0.174&0.093&0.218&0.082&0.208 \\
& Avg 
& {0.031} & {0.113} &
\secondres{0.029}&\secondres{0.107}&\boldres{0.027}&\boldres{0.107}&0.047&0.140&0.120&0.253&0.104&0.218&0.093&0.206&0.062&0.177&0.036&0.126&0.051&0.150&0.071&0.188&0.055&0.166 \\
\midrule

\multirow{5}{*}{\rotatebox{90}{$ETTm2$}}
& 12.5\% 
& 0.019 & 0.079 &
0.018&0.078&0.018&0.080&0.026&0.094&0.108&0.239&0.034&0.127&0.062&0.166&0.056&0.159&0.021&0.088&0.023&0.092&0.133&0.270&0.108&0.228 \\
& 25\% 
& 0.022 & 0.086 &
0.021&0.084&0.020&0.085&0.028&0.099&0.164&0.294&0.042&0.143&0.085&0.196&0.080&0.195&0.024&0.096&0.026&0.101&0.135&0.272&0.136&0.262 \\
& 37.5\% 
& 0.024 & 0.092 &
0.024&0.091&0.023&0.091&0.030&0.104&0.237&0.356&0.051&0.159&0.106&0.222&0.110&0.231&0.027&0.103&0.030&0.108&0.155&0.293&0.175&0.300 \\
& 50\% 
& 0.028 & 0.100 &
0.027&0.098&0.026&0.098&0.034&0.110&0.323&0.421&0.059&0.174&0.131&0.247&0.156&0.276&0.030&0.108&0.035&0.119&0.200&0.333&0.211&0.329 \\
& Avg 
& \secondres{0.023} & {0.089} &
\secondres{0.023}&\secondres{0.087}&\boldres{0.022}&\boldres{0.088}&0.029&0.102&0.208&0.327&0.046&0.151&0.096&0.208&0.101&0.215&0.026&0.099&0.029&0.105&0.156&0.292&0.157&0.280 \\

\bottomrule

\end{tabular}
}
\end{small}
\end{center}
\end{table}

%% file: tables/anomaly_full.tex
\begin{table}[h]
\caption{\textbf{Full Results for the Anomaly Detection.} P represents Precision, R represents Recall, and F1 represents the F1-score. A higher value indicates better performance. \boldres{Red}: the best.}
\label{tab:anomaly_full}
\vskip 0.15in
\begin{center}
\begin{small}
\scalebox{0.75}{
\begin{tabular}{c|ccc|ccc|ccc|ccc|ccc|c}
\toprule

Methods &
\multicolumn{3}{c|}{SMD} & \multicolumn{3}{c|}{MSL} & \multicolumn{3}{c|}{SMAP}& \multicolumn{3}{c|}{SWaT} &\multicolumn{3}{c|}{PSM} & Avg F1 \\
Metrics&P&R&F1&P&R&F1&P&R&F1&P&R&F1&P&R&F1&\%  \\

\midrule
\method{}&0.8787&0.8309&85.42&81.58&82.95&82.26&85.40&\boldres{71.84}&\boldres{78.04}&\boldres{97.90}&92.53&\boldres{94.57}&98.47&\boldres{95.94}&97.19&\boldres{87.51}\\
GPT4TS&88.89&\boldres{84.98}&\boldres{86.89}&82.00&82.91&{82.45}&90.60&60.95&72.88&92.20&96.34&94.23&98.62&95.68&97.13&86.72\\
TimesNet&87.91&81.54&84.61&\boldres{89.54}&75.36&81.84&90.14&56.40&69.39&90.75&95.40&93.02&98.51&95.20&\boldres{97.34}&85.24\\
PatchTST&87.26&82.14&84.62&88.34&70.96&78.70&90.64&55.46&68.82&91.10&80.94&85.72&98.84&93.47&96.08&82.79\\
ETSformer&87.44&79.23&83.13&85.13&84.93&\boldres{85.03}&92.25&55.75&69.50&90.02&80.36&84.91&\boldres{99.31}&85.28&91.76&82.87\\
FEDformer&87.95&82.39&85.08&77.14&80.07&78.57&90.47&58.10&70.76&90.17&96.42&93.19&97.31&97.16&97.23&84.97\\
LightTS&87.10&78.42&82.53&82.40&75.78&78.95&\boldres{92.58}&55.27&69.21&91.98&94.72&93.33&98.37&95.97&97.15&84.23 \\
DLinear&83.62&71.52&77.10&84.34&85.42&84.88&92.32&55.41&69.26&80.91&95.30&87.52&98.28&89.26&93.55&82.46 \\
Stationary&88.33&81.21&84.62&68.55&\boldres{89.14}&77.50&89.37&59.02&71.09&68.03&96.75&79.88&97.82&96.76&97.29&82.08 \\
Autoformer&88.06&82.35&85.11&77.27&80.92&79.05&90.40&58.62&71.12&89.85&95.81&92.74&99.08&88.15&93.29&84.26 \\
Pyraformer&85.61&80.61&83.04&83.81&85.93&84.86&92.54&57.71&71.09&87.92&96.00&91.78&71.67&96.02&82.08&82.57 \\
Anomaly Transformer&88.91&82.23&85.49&79.61&87.37&83.31&91.85&58.11&71.18&72.51&\boldres{97.32}&83.10&68.35&94.72&79.40&80.50 \\
Informer&86.60&77.23&81.65&81.77&86.48&84.06&90.11&57.13&69.92&70.29&96.75&81.43&64.27&96.33&77.10&78.83 \\
Reformer&82.58&69.24&75.32&85.51&83.31&84.40&90.91&57.44&70.40&72.50&96.53&82.80&59.93&95.38&73.61&77.31 \\
LogTransformer&83.46&70.13&76.21&73.05&87.37&79.57&89.15&57.59&69.97&68.67&97.32&80.52&63.06&98.00&76.74&76.60 \\
Transformer&83.58&76.13&79.56&71.57&87.37&78.68&89.37&57.12&69.70&68.84&96.53&80.37&62.75&96.56&76.07&76.88 \\

\bottomrule
\end{tabular}
}
\end{small}
\end{center}
\vskip -0.1in
\end{table}